\documentclass[sigconf]{acmart}

\usepackage{amsfonts}    
\usepackage{amsmath}
\usepackage{booktabs}   
\usepackage{pifont}

\usepackage{xcolor}
\usepackage{multirow}
\usepackage{bm}
\usepackage{array}
\usepackage{color}
\usepackage{colortbl}
\usepackage{textcomp}
\usepackage{stfloats}
\usepackage{hyperref}
\usepackage{verbatim}
\usepackage{graphicx}
\usepackage{booktabs}
\usepackage{subfig}
\usepackage{paralist}

\allowdisplaybreaks 

\usepackage{multirow} 
\usepackage{booktabs} 
\usepackage{arydshln} 

\usepackage{tcolorbox}
\usepackage{adjustbox}
\usepackage{CJKutf8}

\usepackage{pgfplots}
\pgfplotsset{compat=1.17}
\tcbuselibrary{breakable}

\definecolor{red}{RGB}{255,44,0}
\definecolor{ired}{RGB}{229,72,72}
\definecolor{igreen}{RGB}{80,219,144}
\copyrightyear{2024}
\acmYear{2024}
\setcopyright{rightsretained}
\acmConference[MM '24]{Proceedings of the 32nd ACM International Conference on Multimedia}{October 28-November 1, 2024}{Melbourne, VIC, Australia}
\acmBooktitle{Proceedings of the 32nd ACM International Conference on Multimedia (MM '24), October 28-November 1, 2024, Melbourne, VIC, Australia}
\acmDOI{10.1145/3664647.3680705}
\acmISBN{979-8-4007-0686-8/24/10}

\newcommand{\customfont}{\linespread{0.7}\selectfont}

\newcommand{\paratitle}[1]{\vspace{1.2ex}\noindent \textbf{#1}}

\makeatletter
\gdef\@copyrightpermission{
  \begin{minipage}{0.3\columnwidth}
   \href{https://creativecommons.org/licenses/by/4.0/}{\includegraphics[width=0.90\textwidth]{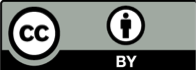}}
  \end{minipage}\hfill
  \begin{minipage}{0.7\columnwidth}
   \href{https://creativecommons.org/licenses/by/4.0/}{This work is licensed under a Creative Commons Attribution International 4.0 License.}
  \end{minipage}
  \vspace{5pt}
}
\makeatother

\begin{document}

\title[Multimodal Conversational Aspect-based Sentiment Analysis]{\texttt{PanoSent}: A Panoptic Sextuple Extraction Benchmark for Multimodal Conversational Aspect-based Sentiment Analysis}


\author{Meng Luo}
\affiliation{%
  \institution{National University of Singapore}
  \city{Singapore}
  \country{Singapore}}
\email{mluo@u.nus.edu}

\author{Hao Fei}
\authornote{Hao Fei is the corresponding author.}
\affiliation{%
  \institution{National University of Singapore}
  \city{Singapore}
  \country{Singapore}}
\email{haofei37@nus.edu.sg}

\author{Bobo Li}
\affiliation{%
  \institution{Wuhan University}
  \city{Wuhan}
  \country{China}}
\email{boboli@whu.edu.cn}

\author{Shengqiong Wu}
\affiliation{%
  \institution{National University of Singapore}
  \city{Singapore}
  \country{Singapore}}
\email{swu@u.nus.edu}

\author{Qian Liu}
\affiliation{%
  \institution{The University of Auckland}
  \city{Auckland}
  \country{New Zealand}}
\email{liu.qian@auckland.ac.nz}

\author{Soujanya Poria}
\affiliation{%
  \institution{Singapore University of Technology and Design}
  \city{Singapore}
  \country{Singapore}}
\email{sporia@sutd.edu.sg}

\author{Erik Cambria}
\affiliation{%
  \institution{Nanyang Technological University}
  \city{Singapore}
  \country{Singapore}}
\email{cambria@ntu.edu.sg}

\author{Mong-Li Lee}
\affiliation{%
  \institution{National University of Singapore}
  \city{Singapore}
  \country{Singapore}}
\email{dcsleeml@nus.edu.sg}

\author{Wynne Hsu}
\affiliation{%
  \institution{National University of Singapore}
  \city{Singapore}
  \country{Singapore}}
\email{dcshsuw@nus.edu.sg}

\renewcommand{\shortauthors}{Meng Luo et al.}


\begin{abstract}
While existing Aspect-based Sentiment Analysis (ABSA) has received extensive effort and advancement, there are still gaps in 
defining a more holistic research target seamlessly integrating multimodality, conversation context, fine-granularity, and also covering the changing sentiment dynamics as well as cognitive causal rationales.
This paper bridges the gaps by introducing a multimodal conversational ABSA, where two novel subtasks are proposed: 
1) \textbf{Panoptic Sentiment Sextuple Extraction}, panoramically recognizing \emph{holder, target, aspect, opinion, sentiment, rationale} from multi-turn multi-party multimodal dialogue.
2) \textbf{Sentiment Flipping Analysis}, detecting the dynamic sentiment transformation throughout the conversation with the causal reasons.
To benchmark the tasks, we construct PanoSent, a dataset annotated both manually and automatically, featuring high quality, large scale, multimodality, multilingualism, multi-scenarios, and covering both implicit\&explicit sentiment elements.
To effectively address the tasks, we devise a novel Chain-of-Sentiment reasoning framework, together with a novel multimodal large language model (namely Sentica) and a paraphrase-based verification mechanism.
Extensive evaluations demonstrate the superiority of our methods over strong baselines, validating the efficacy of all our proposed methods.
The work is expected to open up a new era for the ABSA community, and thus all our codes and data are open at \url{https://PanoSent.github.io/}.
\end{abstract}





\maketitle

\vspace{-2mm}
\section{Introduction}
The quest for human-level artificial intelligence encompasses not only possessing intelligence but also understanding human emotions, thus propelling sentiment analysis and opinion mining to become the key area of research focus.
Through decades of research, sentiment analysis has seen significant developments across various dimensions and aspects~\cite{nasukawa2003sentiment,medhat2014sentiment,cambria2024senticnet}.
The field has evolved from traditional coarse-grained analysis, such as document and sentence-level analysis~\cite{turney2002thumbs,yu2003towards}, to fine-grained one (e.g., ABSA)~\cite{schouten2015survey,nazir2020issues,zhang2022survey}, incorporating a wide array of emotional elements and evolving to extract different sentiment tuples, including the extraction of \emph{targets, aspects, opinions}, and \emph{sentiments}. 
Moreover, the sentiment analysis scope has broadened from purely textual content to multimodal content such as images and videos~\cite{yu2019entity,li2022fine,ji2022knowing,li-etal-2023-qap,fei2024empathyear,luo2024nus}. 
Such expansion recognizes that in real-world scenarios, users often convey their opinions and emotions more accurately through diverse multimedia, providing additional information beyond text, such as micro-expressions, tone of voice, and other cues. 
Additionally, research has expanded beyond single-text scenarios to consider more complex conversational contexts~\cite{zheng-etal-2023-facial,li2023revisiting}, where individuals frequently engage in multi-turn, multi-party discussions on social media platforms (e.g., Twitter, Facebook) about services, products, sports, etc.

\begin{figure}[t]
 \centering
 \includegraphics[width=\columnwidth]{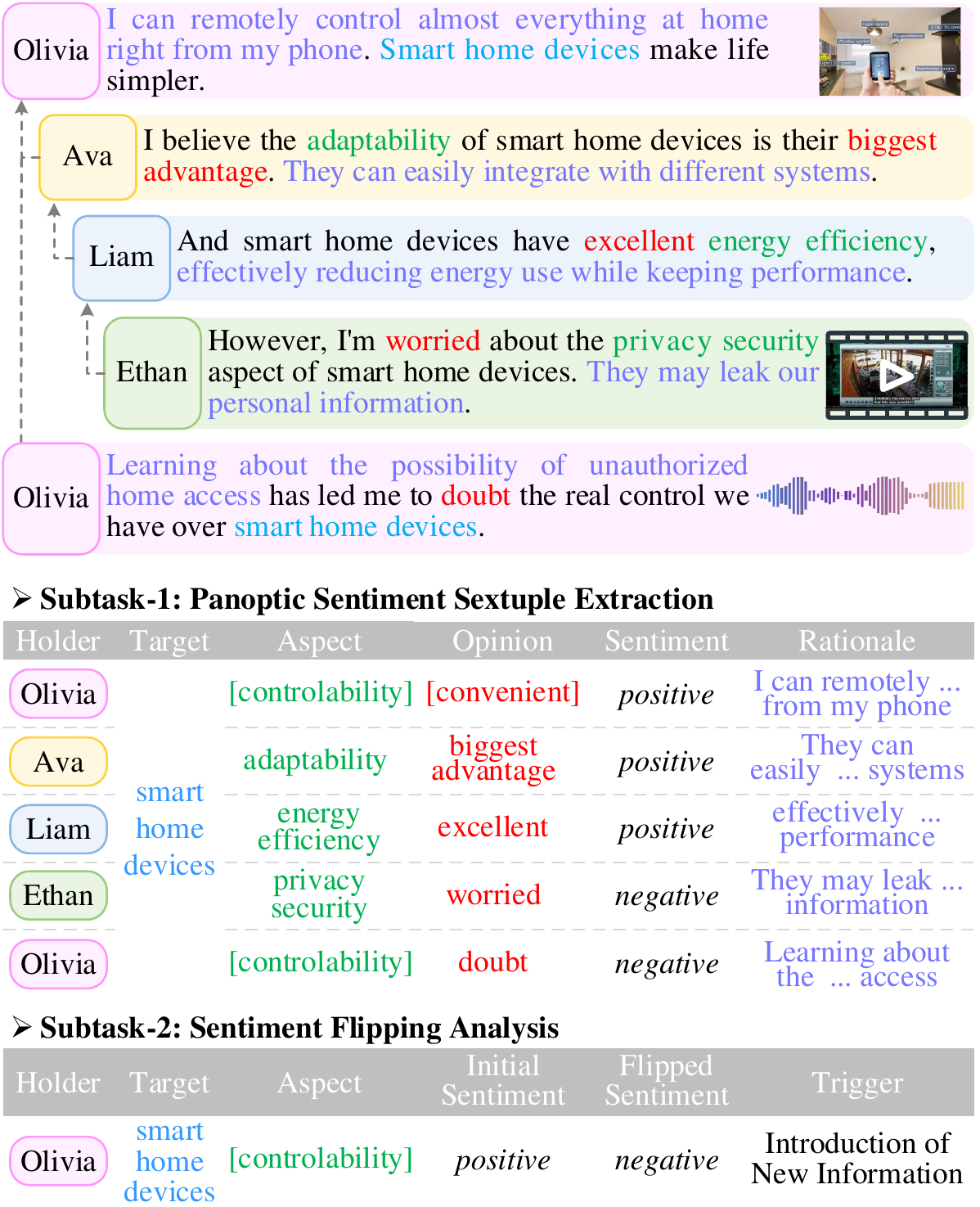}
 \vspace{-7mm}
 \caption{Illustration of the PanoSent benchmark. In \texttt{[*]} are the implicit elements that should be inferred from contexts.
 }
 \vspace{-6mm}
 \label{intro}
\end{figure}

Despite significant progress, current research definitions of sentiment analysis are still not comprehensive enough to offer a complete and detailed emotional picture, primarily due to several issues. 
\textbf{First}, there is a lack of an integrated definition that combines fine-grained analysis, multimodality, and conversational scenarios. 
In real-life applications, such as on social media and forums, these aspects often need to be considered together. 
However, existing studies either lack detailed analysis in multimodal sentiment analysis definitions~\cite{soleymani2017survey,majumder2018multimodal} or miss multimodal modeling in conversational ABSA~\cite{hu2022mm,li-etal-2023-diaasq}. 
The most complete text-based ABSA definitions still do not fully cover or finely detail the granularity of emotional elements. 
\textbf{Second}, current sentiment analysis definitions only consider identifying fixed static emotional polarities~\cite{chakraborty2020survey,birjali2021comprehensive}, neglecting the dynamic nature of emotions that change over time or due to various factors. 
For example, a person's original opinion in a social media conversation may change after being exposed to new information or viewpoints from other speakers. 
\textbf{Third}, and most critically, existing work has not thoroughly analyzed or identified the causal reasons and intentions behind sentiments~\cite{mehta2020review,nandwani2021review}. 
The arousal and change of human emotions have specific triggers, and failing to understand the causal rationale behind emotions from a cognitive perspective means that human-level emotional intelligence has not been fundamentally achieved. 
Overall, providing a more comprehensive sentiment analysis definition could significantly enhance the practical value of this task, e.g., developing smarter voice assistants, better clinical diagnostic and treatment aids, and more anthropomorphic customer service systems.

\definecolor{lightblue}{rgb}{0.93, 0.95, 1.0} 
\definecolor{lightgreen}{rgb}{0.90, 1.0, 0.90} 

\newcolumntype{C}[1]{>{\centering\arraybackslash\hspace{0pt}}p{#1}}

\begin{table*}[!t]
\fontsize{8}{8}\selectfont
\setlength{\tabcolsep}{1.4mm}
\caption{Summary of existing popular benchmarks of sentiment analysis (representatively summarized, not fully covered).
}
\label{tab:benchmark-summary}
\vspace{-3mm}
\rowcolors{2}{lightblue}{white} 
\resizebox{\textwidth}{!}{ 
\begin{tabular}{lcllccC{1.4cm}C{1.4cm}}
  \toprule
  \bf Benchmark&\bf Granularity&\bf Sentiment Picture&\bf Modality&\bf Scenario&\bf Language&\bf Causal\quad Rationale&\bf Sentiment Change\\
  \midrule
  CR~\cite{blitzer-etal-2007-biographies} & Coarse & Sentiment & Text & Sentence & EN &\multicolumn{1}{c}{\textcolor{ired}{\ding{55}}} & \textcolor{ired}{\ding{55}}\\
  Yelp~\cite{tang-etal-2015-document} & Coarse& Sentiment & Text & Document & EN &\multicolumn{1}{c}{\textcolor{ired}{\ding{55}}} & \textcolor{ired}{\ding{55}}\\
  SemEval~\cite{pontiki-etal-2014-semeval} & Fine & Target, Aspect, Sentiment & Text & Sentence & EN &\multicolumn{1}{c}{\textcolor{ired}{\ding{55}}} & \textcolor{ired}{\ding{55}}\\
  TOWE~\cite{fan-etal-2019-target} & Fine & Aspect, Opinion & Text & Sentence & EN &\multicolumn{1}{c}{\textcolor{ired}{\ding{55}}} & \textcolor{ired}{\ding{55}}\\
  ACOS~\cite{cai-etal-2021-aspect} & Fine & Target, Aspect, Opinion, Sentiment & Text & Sentence & EN &\multicolumn{1}{c}{\textcolor{ired}{\ding{55}}} & \textcolor{ired}{\ding{55}}\\
  ASTE~\cite{peng2020knowing} & Fine &Aspect, Opinion, Sentiment & Text & Sentence & EN &\multicolumn{1}{c}{\textcolor{ired}{\ding{55}}} & \textcolor{ired}{\ding{55}}\\
  DiaASQ~\cite{li-etal-2023-diaasq} & Fine &Target, Aspect, Opinion, Sentiment & Text & Dialogue & EN, ZH &\multicolumn{1}{c}{\textcolor{ired}{\ding{55}}} & \textcolor{ired}{\ding{55}}\\
  Twitter2015~\cite{ma-etal-2017-detect} & Fine & Target, Sentiment & Text, Image & Sentence & EN &\multicolumn{1}{c}{\textcolor{ired}{\ding{55}}} & \textcolor{ired}{\ding{55}}\\
  CMU-MOSEI~\cite{zadeh2018multimodal} & Coarse & Sentiment & Text, Audio, Video & Sentence & EN &\multicolumn{1}{c}{\textcolor{ired}{\ding{55}}} & \textcolor{ired}{\ding{55}}\\
  IEMOCAP~\cite{busso2008iemocap} & Coarse & Sentiment & Text, Audio, Video & Dialogue & EN &\multicolumn{1}{c}{\textcolor{ired}{\ding{55}}} & \textcolor{ired}{\ding{55}}\\
  MELD~\cite{poria-etal-2019-meld} & Coarse & Sentiment & Text, Audio, Video & Dialogue & EN &\multicolumn{1}{c}{\textcolor{ired}{\ding{55}}} & \textcolor{ired}{\ding{55}}\\
  M3ED~\cite{zhao-etal-2022-m3ed} & Coarse & Sentiment & Text, Audio, Video & Dialogue & ZH &\multicolumn{1}{c}{\textcolor{ired}{\ding{55}}} & \textcolor{ired}{\ding{55}}\\
  \hdashline
  \rowcolor{white}
  \bf{\texttt{PanoSent}} & Fine & Holder, Target, Aspect, Opinion, Sentiment, Rationale & Text, Image, Audio, Video & Dialogue & EN, ZH, SP &\multicolumn{1}{c}{\textcolor{igreen}{\ding{51}}} & \textcolor{igreen}{\ding{51}}\\
 \bottomrule
\end{tabular}
} 
\vspace{-2mm}
\end{table*}

To fill these gaps, this paper proposes \textbf{Multimodal Conversational Aspect-based Sentiment Analysis},
where we aim to provide a more comprehensive and holistic ABSA definition that includes both \textbf{Panoptic Sentiment Sextuple Extraction} (subtask-I) and \textbf{Sentiment Flipping Analysis} (subtask-II), as exemplified in Figure~\ref{intro}. 
Our focus is on conversational scenarios covering the four most common modalities for emotional expression in daily life, i.e., \emph{text, image, audio, video}. 
On the one hand, we extend the current ABSA quadruple extraction definition to sextuple extraction, including \emph{holder, target, aspect, opinion, sentiment}, and \emph{rationale}, fully covering finer-grained emotional elements to offer a panoramic view of sentiment. 
On the other hand, we define a task to monitor the dynamic sentiment change 
towards the same target and aspect by the same holder throughout the conversation, and also identify the trigger reasons behind these flipped sentiments. 
For both sextuple extraction and sentiment change identification, we also emphasize discerning the underlying causal rationale or trigger, striving to not only know how but also why from a cognition perspective.

To benchmark the novel task, we accordingly construct a large-scale high-quality dataset, \textbf{PanoSent}.
PanoSent covers more than 100 common domains and scenarios, which, based on multi-turn and multi-party conversational contexts, the sentiment elements within a sextuple may cross utterances. 
To mimic real human emotional expression habits, where 1) elements can originate from both textual and non-textual (audio or visual) modalities, and 2) emotions may be expressed implicitly, the data covers both implicit and explicit sentiment elements. 
To ensure the benchmark generalizability, the dataset includes three mainstream languages: English, Chinese, and Spanish. 
We collect the data from real-world sources, carefully annotated manually.
To enlarge the quantity, we further automatically synthesize the dataset via OpenAI GPT-4~\cite{achiam2023gpt} with multimodal retrieval.
Strict human inspection and cross-validation ensure high-quality standards.
In total, we obtain 10,000 annotated dialogues for PanoSent.

Compared to existing ABSA tasks, the new task proposed in this work poses greater challenges, such as the need to understand complex conversational contexts and flexibly extract features from various modalities, especially discerning causal reasons at a cognitive level. 
Considering the recent great successes of Multimodal Large Language Models (MLLMs) in powerful semantic understanding across multiple modalities~\cite{lian2023gpt,wu24next,liu2024visual,fei2024enhancing}, we construct a backbone MLLM system, \textbf{Sentica}, for encoding and understanding multimodal conversational content.
Inspired by the human process of sentiment analysis, we further develop a Chain-of-Sentiment (CoS) reasoning framework for a high-performing task solution, which, based on the Chain-of-Thought~\cite{wei2022chain} idea, breaks down the task into four progressive reasoning steps, from simpler to more complex. 
The system allows to more effectively extract the elements of the sentiment sextuple and identify flipped sentiments step by step, while simultaneously inducing the corresponding rationale and triggers.
A paraphrase-based verification (PpV) mechanism enhances the robustness of the CoS reasoning process. Evaluations on the PanoSent dataset across multiple subtasks and languages show our method outperforms strong LLM-based baselines, validating Sentica, CoS, and PpV. Comprehensive analyses are included for clarity.

In summary, this work makes three significant contributions:

\begin{compactitem}
  \item For the first time, we thoroughly upgrade ABSA with a more comprehensive definition at the cognitive level, Multimodal Conversational Aspect-based Sentiment Analysis, introducing Panoptic Sentiment Sextuple Extraction and Sentiment Flipping Analysis tasks, achieving the ultimate form of sentiment analysis within the community.
  
  \item We contribute a large-scale, high-quality benchmark dataset, PanoSent, featuring multiple aspects: conversational contexts, multimodality, multilingualism, and multidomain.
  
  \item We propose an advanced reasoning framework, the Chain-of-Sentiment, based on our Sentica MLLM, achieving high task performance and providing a strong baseline for subsequent research on PanoSent.

\end{compactitem}

\section{Related Work}
This work majorly focuses on the track of ABSA~\cite{zhang-etal-2021-aspect-sentiment,chen-etal-2021-semantic}. 
ABSA has evolved from its initial objective of identifying sentiment polarity to more complex tasks such as recognizing targets, aspects, and opinions~\cite{liu-etal-2015-fine,jiang-etal-2011-target,liang2022aspect}. 
The complexity of ABSA tasks has increased with the introduction of combinations of these elements, ranging from paired extraction~\cite{chen-etal-2020-synchronous,wu-etal-2020-grid} to triplet~\cite{peng2020knowing,mao2021joint} and quadruple extractions~\cite{cai-etal-2021-aspect,li-etal-2023-diaasq}. 
Concurrently, multimodal SA~\cite{hu2021mmgcn}, a pivotal topic within the multimodal research community~\cite{YangYNGX23,fei2024dysen,fei2023scene,wu2023imagine}, has garnered increasing attention, incorporating modalities beyond text, such as images, audios, and videos. 
The trend in multimodal sentiment analysis has shifted from coarse-grained to fine-grained.
The proposed methods mainly focus on exploring feature extraction and fusion from diverse modal inputs~\cite{yu2019entity,hu-etal-2022-unimse,wu2023information,LingYX22,fei2024enhancing,zhao2023constructing}, relying on additional structured knowledge~\cite{fei2022matching,fei2023scene}. 
Furthermore, in terms of application scenarios, there has been a shift from analyzing single pieces of text to engaging in multi-turn, multi-party dialogues~\cite{zhang-li-2023-cross,zhang-etal-2023-dualgats}, aiming to recognize emotions within dialogues to better align with real-world applications. 
Subsequently, dialogue sentiment analysis has gradually evolved into dialogue ABSA~\cite{li-etal-2023-diaasq}, incorporating non-textual modalities in the analysis.

However, we find that current ABSA benchmarks still lack a combined perspective and comprehensive definition across granularity, multimodality, and dialogue contexts. 
For instance, there is an absence of benchmarks for fine-grained sentiment analysis in multimodal dialogue scenarios~\cite{nazir2020issues,zhang2022survey}. 
Regarding granularity, there is potential to go beyond the four elements of target, aspect, opinion, and sentiment, to include the consideration of the sentiment holder, which also plays a pivotal role in a dialogue context.

Moreover, previous research has not fully leveraged the role of multimodality in ABSA. 
In most cases, multimodal information is merely considered as supplementary clues to assist in determining opinions or sentiments~\cite{marrese-taylor-etal-2020-multi,shi-huang-2023-multiemo}, with most of the other elements (e.g., targets, aspects) coming from texts. 
However, we argue that multimodality can also serve as a crucial source of information for the implicit identification of all elements more than sentiment. 
For example, a `cellphone' may not be mentioned in the utterance, but the image showing a phone might feature it as the `target'12 element.
Beyond that, two other key aspects have not been sufficiently addressed in the existing ABSA. 
First, the dynamic nature of sentiments, especially within the context of dialogues, has not been explored. 
Second, the cognitive causes and intentions behind sentiments have been overlooked. 
In response, this work introduces a new benchmark, \texttt{PanoSent}, aiming to bridge all the above gaps, and provide a platform for the next phase of more comprehensive and in-depth ABSA research. 
Table~\ref{tab:benchmark-summary} summarizes the key differences between ours and existing benchmarks.

Beyond contributing new data, we also propose an advanced methodology for this benchmark.
We take full advantage of the significant success of existing MLLM~\cite{yin2023survey,zhang2024mm,fei2024vitron,wu2024towards} in understanding multimodal data. 
To address the challenges posed by the new tasks, which rely on cognitive-level reasoning, we introduce a novel reasoning framework, CoS. 
Inspired by the existing CoT strategy, which breaks down the problem into smaller chained steps for step-by-step resolution~\cite{wei2022chain,fei2023reasoning}, we decompose the two tasks in PanoSent, significantly enhancing the task-solving efficacy. 
Overall, our new benchmark data and methods are poised to open up a new era for the ABSA community.

\vspace{-2mm}
\section{Task Definition}
We formally give the definitions of two subtasks, which also are illustrated in Figure~\ref{intro} with specific examples.

\vspace{-1mm}
\paratitle{Subtask-I: Panoptic Sentiment Sextuple Extraction.}
Given a dialogue $D = \{u_1, \ldots, u_n\}$ with the replying structure $\{(u_i, u_j), \ldots \}$ (i.e., $u_i$ replies to $u_j$),
the task is to extract all sextuples $(h, t, a, o, s, r)$.
Each utterance $u_i = \{w_1, \ldots, w_{m_i}\}$ contains $m_i$ words in the text (denoted as $I^t$), occasionally with associated non-text information piece, i.e., image ($I^i$), audio ($I^a$), video ($I^v$).
The elements $h$ (holder), $t$ (target), $a$ (aspect), $o$ (opinion), and $r$ (rationale) can be either the continuous text spans explicitly mentioned in utterances, or implicitly inferred from contexts or non-text modalities.
$s$ represents the sentiment category (positive, negative, or neutral).

\vspace{-1mm}
\paratitle{Subtask-II: Sentiment Flipping Analysis.}
Given input $D$, the same as in subtask-I, the task detects all sextuples $(h, t, a, \zeta, \phi, \tau)$.
Here, $h$, $t$, and $a$ denote the holder, target, and aspect, consistent with the definitions in subtask-I.
$\zeta$ and $\phi$ represent the initial and flipped sentiments, respectively, highlighting the dynamic change in sentiment by the same speaker towards the same aspect of the same target.
$\tau$ refers to a trigger that induces the sentiment transition, which is a pre-defined label among four categories: 1) \textit{introduction of new information}, 2) \textit{logical argumentation}, 3) \textit{participant feedback and interaction}, and 4) \textit{personal experience and self-reflection}.
Since subtask-II shares multiple elements with subtask-I, it is natural to detect the flipping based on the results from subtask-I to minimize redundancy.

\vspace{-2mm}
\section{New benchmark: \texttt{PanoSent}}

Here we elaborate on the construction of the new dataset for multimodal conversational ABSA, as well as its key characteristics.

\vspace{-3mm}
\subsection{Dataset Construction}

\vspace{-2mm}
\paratitle{Constructing via Human Annotation.}
The corpus of dialogues is collected by crawling via publicly available APIs from various social media or forum platforms in different languages, such as Twitter, Facebook, Reddit, Weibo, Xiaohongshu, BeReal, and more.
While the majority of these dialogues are text-based, some also include multimodal interactions.
Then, we conduct a rigorous screening process (via both manual inspection and automated filters, e.g., keyword and Toxic-BERT detection\footnote{\url{https://github.com/unitaryai/detoxify}}), to eliminate content (e.g., multimodal information) or instances that are harmful, private or unrelated to the dialogue.
After obtaining a cleansed corpus, we commence the annotation of aspect-based sentiment sextuples.
We stick to the SemEval guidelines~\cite{pontiki-etal-2014-semeval} and customize the annotation manual to accommodate both subtasks of our benchmark.
We recruit annotators, training them according to the manual.
To guarantee reliability, each dialogue is annotated independently by at least three distinct annotators.
After annotation, we calculate the Cohen's Kappa score~\cite{cohen1960coefficient}, achieving a score of \textbf{0.85}, which reflects the high quality of our annotated dataset.
In instances with inconsistent annotations, linguists and native speakers will collaboratively determine the final annotation.
For unresolved ambiguities, the instances will be dropped.

\begin{table}[!t]
\fontsize{9}{10}\selectfont
\setlength{\tabcolsep}{1mm}
\centering
\caption{Main statistics of PanoSent dataset. 
}
\vspace{-3mm}
\resizebox{\columnwidth}{!}{%
\begin{tabular}{@{} llcccccccccccccccc @{}}
\toprule
& & \multicolumn{3}{c}{Dialogue} & \multicolumn{2}{c}{Sextuple} & \multicolumn{5}{c}{Modality} & \multicolumn{2}{c}{Manner} \\
\cmidrule(r){3-5} \cmidrule(lr){6-7} \cmidrule(lr){8-12} \cmidrule(lr){13-14}
& & Dia. & Utt. & Spk. & Sext. & Flip. & Txt. & Img. & Aud. & Vid. & Mix. & Imp. & Exp. \\

\midrule
\multirow{3}{*}{EN}
& Total & 6,000 & 28,822 & 26,831 & 28,464 & 2,136 & 3,360 & 1,320 & 360 & 240 & 720 & 1,680 & 4,320 \\
\cdashline{2-14}
& Real & 2,000 & 9,573 & 8,827 & 9,298 & 694 & 1,102 & 427 & 108 & 70 & 232 & 536 & 1,464 \\
& Synth & 4,000 & 19,249 & 18,004 & 19,166 & 1,442 & 2,258 & 893 & 252 & 170 & 488 & 1,144 & 2,856 \\

\midrule
\multirow{3}{*}{ZH}
& Total & 3,000 & 14,033 & 13,444 & 13,965 & 1,068 & 1,680 & 660 & 180 & 120 & 360 & 840 & 2,160 \\
\cdashline{2-14}
& Real & 1,000 & 4,702 & 4,510 & 4,672 & 360 & 582 & 210 & 63 & 41 & 125 & 289 & 711 \\
& Synth & 2,000 & 9,331 & 8,934 & 9,293 & 708 & 1,098 & 450 & 117 & 79 & 235 & 551 & 1,449 \\

\midrule
\multirow{3}{*}{SP}
& Total & 1,000 & 4,667 & 4,490 & 4,671 & 356 & 560 & 220 & 60 & 40 & 120 & 280 & 720 \\
\cdashline{2-14}
& Real & 333 & 1,547 & 1,488 & 1,551 & 114 & 181 & 72 & 18 & 12 & 35 & 90 & 243 \\
& Synth & 667 & 3,120 & 3,002 & 3,120 & 242 & 379 & 148 & 42 & 28 & 75 & 190 & 477 \\

\midrule
& All & 10,000 & 47,522 & 44,765 & 47,100 & 3,560 & 5,600 & 2,200 & 600 & 400 & 1,200 & 2,800 & 7,200 \\
\bottomrule
\end{tabular}
} 
\vspace{-7mm}
\label{tab:data}
\end{table}

\paratitle{Constructing via Auto-Synthesis.}
We find the cost and workload in the above manual annotation process to be significantly high. 
The key issue is that real-world data sources that can provide a sufficient data volume meeting our task definition (to cover various modalities) are very rare.
Hence, we consider automating data synthesis to substantially expand the volume, with the basic idea of `\emph{automatic synthesis + multimodal retrieval}'. 
We first leverage the powerful LLMs for synthesizing dialogues and sextuples.
A considerable amount of existing related work~\cite{peng2023instruction,ding2023enhancing,mukherjee2023orca} has already demonstrated that OpenAI's GPT-4 can generate data of very high quality that almost perfectly matches the real distribution.
Specifically, following the prior practices~\cite{xu2023wizardlm,ding2023enhancing}, we prepare template prompts to guide GPT-4 to generate pseudo-dialogues, along with sextuple and flipping annotations.
Besides, for a portion of dialogue utterances, we also instruct GPT-4 to create appropriate captions as the image, audio, and video placeholders, according to the contexts.

With the annotated dialogues, we now use the captions to retrieve the piece of information in the corresponding modality (image, audio or video) from the external multimodal databases, with only the top-10 retrieved candidates kept.
Specifically, we consider multiple large-scale databases, including COCO~\cite{lin2014microsoft}, Flickr30k~\cite{young-etal-2014-image}, AudioSet~\cite{gemmeke2017audio}, WaveText5K~\cite{deshmukh2022audio}, and WebVid~\cite{bain2021frozen}, etc.
Also we consider direct retrieval from the Google search engine, to ensure comprehensive coverage.
For the associated multimodal contents, three annotators will assign a ranking score (1-10) to the 10 candidates, which
are further ranked via their averaged scores, and the highest-scored one is elected as the determined multimodal information piece.
Finally, each synthesized dialogue, the annotations of two subtasks, and the multimodal contents will be thoroughly examined by at least two workers.
All the possibly problematic instances will be dropped.
We also calculate the Cohen's Kappa score across workers, achieving a score of \textbf{0.82}, ensuring a high consistency of the synthesized annotations.

\vspace{-2mm}
\subsection{Data Insights}
\label{Data Insights}

We select a portion of the real data to serve as developing and testing sets, while the remainder of the real data and all the synthesized data are used as the training set. 
Ultimately, the ratio of the train/dev/test sets for each language is 8:1:1.
Following we briefly summarize the key characteristics and highlights of our PanoSent dataset.

\vspace{-1mm}
\paratitle{Panoptic Fine-grained Sentiment Definition.}
In contrast to existing ABSA datasets, such as TOWE~\cite{fan-etal-2019-target}, ASTE~\cite{peng2020knowing}, and DiaASQ~\cite{li-etal-2023-diaasq}, PanoSent dataset encompasses the most comprehensive elements, featuring six key items for ABSA.

\vspace{-1mm}
\paratitle{Cognitive Causal Rationale.}
We for the first time introduce the rationale element in ABSA, enhancing the definition by providing deeper insights into the motivations behind sentiments, allowing an interpretable sentiment understanding at a cognitive level.

\vspace{-1mm}
\paratitle{Dynamic Sentiment Flipping.}
Going beyond the traditional ABSA benchmark, PanoSent pioneers the examination of sentiment flips, studying the dynamics nature of ABSA.

\vspace{-1mm}
\paratitle{Multi-scenario.}
PanoSent takes the dialogue as the context backbone, covering 10 main real-life domains across over 100 sub-domains, ensuring an extensive diversity that supports research into sentiment analysis from various perspectives.

\vspace{-1mm}
\paratitle{Multimodality.}
Beyond textual content (56\%), PanoSent comprises three other modalities of information, including images (22\%), audio (6\%), video (4\%), and mixed modalities (12\%).

\vspace{-1mm}
\paratitle{Multilingualism.}
PanoSent covers three mainstream languages, English (60\%), Chinese (30\%), and Spanish (10\%), allowing a cross-lingual study of ABSA.

\vspace{-1mm}
\paratitle{Implicit ABSA.}
Our dataset fully supports implicit ABSA, thereby elevating the challenges. 
While most of the sextuples are explicitly mentioned in the utterance text, 28\% of the dialogues contain elements that need to be implicitly inferred from contexts or various modality information.

\vspace{-1mm}
\paratitle{High-quality and Large-scale.}
Through careful manual annotation and cross-validation, we ensure the high quality of PanoSent.
By employing automated synthesis, we significantly expand the scale of the dataset without compromising its quality, resulting in a total of 10,000 dialogue instances and 47,100 sextuples. The statistics are presented in Table~\ref{tab:data}.

\begin{figure}[!t]
 \centering
 \includegraphics[width=\columnwidth]{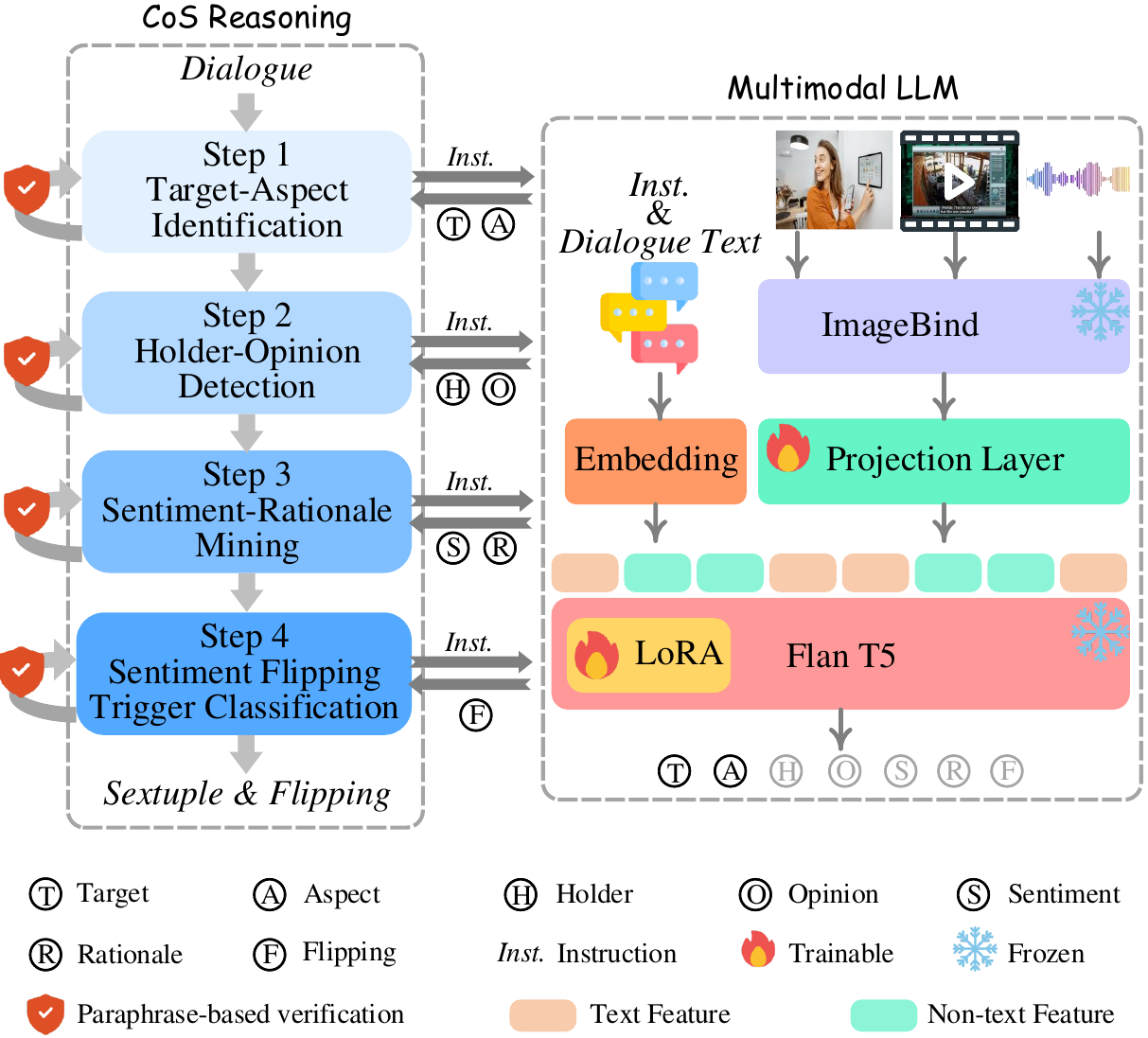}
 \caption{Schematic overview of our \textbf{Sentica} MLLM.}
 \label{framework}
\vspace{-4mm}

\end{figure}

\section{Methodology}

The two tasks in PanoSent encompass non-trivial challenges, e.g., complex conversational context understanding, multimodal feature extracting, and cognitive-level ABSA reasoning. 
To address these, we propose a comprehensive solution. 
Below, we detail the models proposed, the reasoning framework, the verification mechanism, and the learning approach.

\vspace{-2mm}
\subsection{Multimodal LLM Backbone}

Currently, LLMs demonstrate remarkable capabilities in understanding language semantics. 
Correspondingly, MLLMs have been developed, exhibiting powerful abilities to comprehend multimodal data~\cite{li2023mimic}. 
Building on the success of MLLMs, we consider leveraging them to help solve our task, where a thorough understanding of multimodal information is required. 
To this end, we develop a novel MLLM, \textbf{Sentica}, as presented in Figure~\ref{framework}.
We adopt the Flan-T5 (XXL)~\cite{chung2022scaling} as the core LLM for semantics understanding and decision-making.
For non-text inputs, we use multimodal models to encode signals into LLM-understandable representations. We use ImageBind as the unified encoder for all three non-text modalities due to its strong capabilities, followed by a linear layer that connects ImageBind~\cite{girdhar2023imagebind} to the LLM for representation projection.

\subsection{CoS Reasoning Framework}
Resolving two tasks, Panoptic Sentiment Sextuple Extraction and Sentiment Flipping Analysis, is challenging, not only due to the complex task definitions but also the cognitive-level requirement on the causal rationale and trigger detection.
Inspired by the recent Chain-of-Thought (CoT) reasoning paradigm~\cite{wei2022chain}, here we also consider a human-like process of sentiment understanding and propose a Chain-of-Sentiment (CoS) reasoning framework.
Previous ABSA studies~\cite{fei2022inheriting} reveal that various ABSA elements can play hierarchical roles in depicting the overall sentiment puzzle.
For example, the opinion should be detected before determining the sentiment polarity; likewise, identifying the target and aspect has a higher priority over recognizing the opinion.
Thus, our main idea is that we deconstruct the two subtasks into four progressive, chained reasoning steps, from simpler to more complex.
Using the capability of Sentica, solving each step incrementally accumulates key clues and insights for the follow-up steps.
Figure~\ref{framework} also illustrates how the CoS reasoning works with Sentica.

$ $

\paratitle{Step 1: Target-Aspect Identification.}
Given input dialogue \(D\) possibly with multimodal signals and with specific instruction \(P_1\), the initial step aims to prompt Sentica to identify all the possible \textbf{targets} and their specific \textbf{aspects} discussed within the dialogue, i.e., \(\{(t_i, a_i)\}\). 

\vspace{-2mm}
\begin{tcolorbox}[breakable, fontupper=\customfont]
\vspace{-2mm}
{\small
\textbf{Input Data}: \(D\) \\
\textbf{Instruction}: Based on the multi-party dialogue and its accompanying multimodal data, please identify all possible targets and their specific aspects mentioned in the dialogue. 
Extract each target and aspect explicitly from the utterance text spans, or infer them implicitly via your understanding of the input data. 
Ensure each identified target is paired with its aspect(s), forming target-aspect pairs.\\

\textbf{\color{blue}{Expected Output}}: (target, aspect)$_1$, (target, aspect)$_2$, $\cdots$ \\
}
\vspace{-4mm}
\end{tcolorbox}
\vspace{-2mm}
This step can be formulated as:
\begin{equation}
\setlength\abovedisplayskip{3pt}
\setlength\belowdisplayskip{3pt}
  \{(t_i, a_i)\} \gets f_1(D | P_1) \,.
\end{equation}

\paratitle{Step 2: Holder-Opinion Detection.}
The second step is to detect the \textbf{holders} \(h_j\) and their specific \textbf{opinions} \(o_j\), regarding the identified targets and aspects. 
We require Sentica to output a set of quadruples consisting of the holder, target, aspect, and opinion \(\{(h_j, t_i, a_i, o_j)\}\).
After this step, we construct holder-target-aspect-opinion quadruples, which lay the foundation for understanding the further sentiment.

\vspace{-2mm}
\begin{tcolorbox}[breakable, fontupper=\customfont]
\vspace{-2mm}
{\small
\textbf{Input Data}: \(D\), \(\{(t_i, a_i)\}\) \\
\textbf{Instruction}: Based on the dialogue and each target-aspect pair identified previously, please identify the holder (the person who expresses an opinion, normally should be a speaker of certain dialogue utterance) and the opinion, both either directly extracted from the text or inferred from our understanding of the input data.
Formulate your output into `holder-target-aspect-opinion' quadruples, ensuring each element is clearly identified.\\

\textbf{\color{blue}{Expected Output}}: (holder, target, aspect, opinion)$_1$, (holder, target, aspect, opinion)$_2$, $\cdots$ \\
}
\vspace{-4mm}
\end{tcolorbox}
\vspace{-2mm}
This step is formulated as:
\begin{equation}
\setlength\abovedisplayskip{3pt}
\setlength\belowdisplayskip{3pt}
  \{(h_j, t_i, a_i, o_j)\} \gets f_2(D, \{(t_i, a_i)\} | P_2) \,.
\end{equation}

\paratitle{Step 3: Sentiment-Rationale Mining.}
The third step then analyzes the \textbf{sentiment} \(s_k\) with each opinion and identifies the \textbf{rationale} \(r_l\), based on the identified holder-target-aspect-opinion quadruples. 
We ask Sentica to output a set of sextuplets, by further adding sentiment and rationale to the previous quadruples to form \(\{(h_j, t_i, a_i, o_j, s_k, r_l)\}\).
\vspace{-2mm}
\begin{tcolorbox}[breakable, fontupper=\customfont]
\vspace{-2mm}
{\small
\textbf{Input Data}: \(D\), \(\{(h_j, t_i, a_i, o_j)\}\) \\
\textbf{Instruction}: Based on the dialogue and each holder-target-aspect-opinion quadruple identified previously, please identify the sentiment polarity associated with the opinion and analyze the causal rationale behind it. 
The sentiment polarity should be classified as `positive', `neutral', or `negative'. The rationale should be extracted explicitly from the text, or
inferred implicitly via your understanding of the input data.
Formulate your output into `holder-target-aspect-opinion-sentiment-rationale' sextuples, ensuring sentiment polarity is clearly analyzed and the other five elements are clearly identified.\\

\textbf{\color{blue}{Expected Output}}: (holder, target, aspect, opinion, sentiment, rationale)$_1$, $\cdots$ \\
}
\vspace{-4mm}
\end{tcolorbox}
\vspace{-2mm}
We denote this step as:
\begin{equation}
\setlength\abovedisplayskip{3pt}
\setlength\belowdisplayskip{3pt}
  \{(h_j, t_i, a_i, o_j, s_k, r_l)\} \gets f_3(D, \{(h_j, t_i, a_i, o_j)\} | P_3) \,.
\end{equation}

\begin{table*}[!t]
\fontsize{8}{8}\selectfont
\setlength{\tabcolsep}{1.9mm}
\centering
\caption{Main results of Subtask-I, Panoptic Sentiment Sextuple Extraction. `H/T/A/O/R/S' represents Holder, Target, Aspect, Opinion, Rationale, and Sentiment, respectively. 
All the scores are averaged over five runs under different random seeds.
}
\vspace{-3mm}
\begin{tabular}{@{} lcllccccccccccccccc @{}}
\toprule
\multirow{2}{*}{} & \multicolumn{2}{c}{\multirow{2}{*}{\textbf{Model}}} & \multirow{2}{*}{\textbf{PLM}} & \multicolumn{5}{c}{\textbf{Element-wise}} & \multicolumn{4}{c}{\textbf{Pair-wise}} & \multicolumn{2}{c}{\textbf{Sextuple}} \\
\cmidrule(r){5-9} \cmidrule(l){10-13} \cmidrule(l){14-15}
& & & & H & T & A & O & R & T-A & H-O & S-R & O-S & Micro & Iden. \\

\midrule
\multirow{8}{*}{\textbf{EN}} & M1 & DiaASQ & mBERT Base & 69.56 & 58.61 & 52.04 & 44.39 & 22.90 & 33.07 & 33.52 & 18.98 & 40.26 & 13.49 & 19.07\\
& M2& UGF & mT5-XXL & 71.17 & 61.83 & 55.25 & 47.68 & 25.87 & 35.39 & 36.08 & 22.37 & 42.80 & 15.85 & 20.12 \\ 
\cdashline{2-15}

& M3& Unified-IO 2 & Unified-IO 2 7B & 75.82 & 65.81 & 59.50 & 51.57 & 29.03 & 39.41 & 40.36 & 26.16 & 47.03 & 18.95 & 22.03\\
& M4& NExT-GPT & Vicuna 7B & 76.07 & 66.25 & 59.97 & 52.12 & 29.95 & 40.23 & 41.24 & 27.07 & 47.89 & 20.01 & 24.98\\
& M5& Sentica & Flan-T5-XXL & 77.48 & 67.49 & 61.01 & 53.06 & 31.02 & 41.12 & 42.31 & 28.12 & 48.94 & 21.26 & 25.67\\

& M6& Sentica (+CoT) & Flan-T5-XXL & 80.98 & 72.85 & 67.21 & 58.07 & 38.10 & 46.49 & 47.35 & 34.47 & 55.25 & 26.69 & 30.95\\
& M7& Sentica (+CoS) & Flan-T5-XXL & 83.41 & 75.70 & 70.38 & 60.96 & 41.35 & 49.72 & 50.47 & 37.27 & 58.20 & 29.71 & 33.69 \\
\rowcolor{lightblue}
& M8& Sentica (+CoS+PpV) & Flan-T5-XXL & \textbf{84.30} & \textbf{76.51} & \textbf{71.16} & \textbf{62.47} & \textbf{43.23} & \textbf{51.09} & \textbf{52.20} & \textbf{39.50} & \textbf{60.25} & \textbf{32.18} & \textbf{35.72} \\

\midrule
\multirow{5}{*}{\textbf{ZH}} & M9& DiaASQ & mBERT Base & 66.02 & 55.07 & 50.66 & 40.21 & 18.19 & 29.33 & 30.90 & 16.15 & 37.89 & 11.05 & 16.25\\
& M10& UGF & mT5-XXL & 67.81 & 57.86 & 53.72 & 43.15 & 21.17 & 31.71 & 33.49 & 18.63 & 39.88 & 13.70 & 17.09 \\
\cdashline{2-15}
& M11& Sentica & ChatGLM2 6B  & 74.19 & 64.20 & 58.45 & 49.39 & 28.04 & 38.02 & 38.16 & 24.61 & 45.70 & 18.57 & 22.86 \\
& M12& Sentica (+CoT) & ChatGLM2 6B & 77.76 & 68.82 & 64.21 & 54.43 & 34.70 & 42.87 & 43.23 & 30.69 & 51.58 & 23.64 & 27.88\\

\rowcolor{lightblue}
& M13& Sentica (+CoS+PpV) & ChatGLM2 6B & \textbf{80.05} & \textbf{72.29} & \textbf{67.83} & \textbf{58.25} & \textbf{38.96} & \textbf{46.82} & \textbf{48.04} & \textbf{35.78} & \textbf{56.61} & \textbf{28.06} & \textbf{31.91}\\

\midrule
\multirow{5}{*}{\textbf{SP}}& M14 & DiaASQ & mBERT Base & 63.72 & 53.80 & 46.33 & 36.59 & 17.02 & 26.89 & 29.61 & 14.52 & 35.13 & 8.23 & 13.68 \\
& M15 & UGF & mT5-XXL & 65.14 & 55.69 & 49.17 & 39.57 & 19.89 & 29.44 & 31.02 & 16.03 & 37.06 & 11.11 & 14.92\\
\cdashline{2-15}
& M16 & Sentica & Vicuna 7B & 71.61 & 62.02 & 55.83 & 47.02 & 25.73 & 35.77 & 35.83 & 22.17 & 43.04 & 15.97 & 20.12\\
& M17 & Sentica (+CoT) & Vicuna 7B & 74.89 & 66.34 & 61.83 & 51.94 & 32.51 & 40.26 & 40.88 & 28.07 & 48.84 & 21.16 & 25.40 \\
\rowcolor{lightblue}
& M18 & Sentica (+CoS+PpV) & Vicuna 7B & \textbf{77.49} & \textbf{69.85} & \textbf{65.31} & \textbf{55.62} & \textbf{36.66} & \textbf{44.37} & \textbf{45.54} & \textbf{33.39} & \textbf{54.05} & \textbf{25.62} & \textbf{29.54}\\

\bottomrule
\end{tabular}
\vspace{-3mm}
\label{tab:main1}
\end{table*}

\paratitle{Step 4: Sentiment Flipping Trigger Classification.}
With all the sextuplets detected, the final step of discerning sentiment flipping would be much effortless.
Specifically, we prompt Sentica to first summarize any changes (i.e., from an \textbf{initial sentiment} ($\zeta_k$) to a \textbf{flipped sentiment} ($\phi_k$)) in sentiment of same \emph{holder-target-aspect}, and then classify the \textbf{trigger} ($\tau_m$) label for each sentiment flip.

The output is a set of sextuplets: \(\{(h_j, t_i, a_i, \zeta_k, \phi_k, \tau_m)\}\).
\vspace{-2mm}
\begin{tcolorbox}[breakable, fontupper=\customfont]
\vspace{-2mm}
{\small
\textbf{Input Data}: \(D\), \(\{(h_j, t_i, a_i, o_j, s_k, r_l)\}\) \\
\textbf{Instruction}: Based on the dialogue and each holder-target-aspect-opinion-sentiment-rationale sextuple, please identify instances where a sentiment flip occurs for the same holder regarding the specific target-aspect pair. 
Determine the trigger type for these flips from the predefined categories: \textit{introduction of new information}, \textit{logical argumentation}, \textit{participant feedback and interaction}, \textit{personal experience and self-reflection}.
Formulate your output to include the holder, target, aspect, initial sentiment, flipped sentiment, and the trigger type, or state "None" if no flips are identified.\\

\textbf{\color{blue}{Expected Output}}: (holder, target, aspect, initial sentiment, flipped sentiment, trigger type)$_1$, $\cdots$; or "None" \\
}
\vspace{-4mm}
\end{tcolorbox}
\vspace{-2mm}
This step can be marked as:
\begin{equation}
\setlength\abovedisplayskip{3pt}
\setlength\belowdisplayskip{3pt}
{\scriptsize
\left\{
  \begin{array}{ll}
    \text{NONE}, & \text{if no flip} \\ 
    (h, t, a, \zeta, \phi, \tau), & \text{if flip} 
  \end{array} 
\right\}
\gets f_4\left(D, \{(h_j, t_i, a_i, o_j, s_k, r_l)\} \middle| P_4\right).
}
\end{equation}

\vspace{-2mm}
\subsection{Paraphrase-based Verification}

Given that we designed the entire two-task solution as a step-wise process, a potential issue is that CoS could lead to error accumulation. 
For example, an error in the first step could directly impact the outcome of all subsequent steps. 
Therefore, it's crucial to perform verification at every reasoning step. 
Existing work has verified that compared to structured data, LLMs excel more in understanding natural language~\cite{tan2023make,lee2024learning}. 
This implies that having LLMs directly check the correctness of each obtained $k$-tuple is sub-optimal. 
A more intuitive approach is to first convert the structured $k$-tuples into natural language expressions through paraphrasing, effectively creating a claim that conveys the same meaning in a different format. 
Then, let the LLM check whether this claim is in an entailment or contradiction relationship~\cite{kamoi2023wice,sanyal2024minds} with the given dialogue context and information. 
We refer to this as a \emph{Paraphrase-based Verification} (PpV) mechanism. 
If the relationship is one of entailment, the verification is successful, and the process moves on to the next reasoning step. 
If it's a contradiction, the current step is rerun until a reasonable result is yielded. 
This process not only ensures that each reasoning step is built on verified information but also enhances the overall robustness of sentiment analysis, effectively mitigating the negative impact of hallucinations~\cite{huang2023survey,qian2024easy} inherent in LLMs.

\vspace{-3mm}
\subsection{Instruction Tuning}
To empower Sentica with the reasoning capabilities of the CoS framework, we conduct instruction tuning, entailing a three-phase training process.
In the first stage, we enable the LLM to understand multimodal representations bound to images, audios and videos. We consider training directly on existing `text-X' pair datasets (where `X' refers to image, audio, or video), i.e., inputting `X' and having the LLM output the corresponding caption text.

In the second stage, we aim for the LLM to smoothly and accurately execute the sextuple extraction process. We consider using the PanoSent train set as supervised data, wrapping the corresponding instructions to obtain instruction fine-tuning data. 
Then, we train the model on the data to master the response mode for the corresponding inputs and outputs.
The third stage teaches Sentica the PpV pattern. 
Based on the previous instructions, we construct correct verification pairs with an entailment relation. 
Meanwhile, by arbitrarily altering elements of the $k$-tuple, we create contradictory relations in paraphrases as counterexamples, on which we fine-tune Sentica.

\begin{table*}[!t]
\fontsize{8}{8}\selectfont
\setlength{\tabcolsep}{1.9 mm}
\centering
\caption{Results of the Subtask-II, Sentiment Flipping Analysis.}
\vspace{-3mm}
\begin{tabular}{@{} llllccccccccc @{}}
\toprule
&\multirow{2}{*}{\textbf{Model}} & \multicolumn{3}{c}{\textbf{EN}} & \multicolumn{3}{c}{\textbf{ZH}} & \multicolumn{3}{c}{\textbf{SP}} \\
\cmidrule(lr){3-5} \cmidrule(lr){6-8} \cmidrule(lr){9-11}
& & Flip & Trig & Flip-Trig & Flip & Trig & Flip-Trig & Flip & Trig & Flip-Trig\\
\midrule
M1&NExT-GPT & 60.27 & 63.43 & 55.80 & / & / & / & 51.32 & 55.52 & 46.02 \\
M2&Sentica & 63.71 & 66.26 & 58.49 & 58.83 & 62.50 & 52.57 & 55.37 & 59.61 & 50.98 \\
M3&Sentica (+CoT) & 65.53 & 69.30 & 61.99 & 61.79 & 65.70 & 58.04 & 58.31 & 62.57 & 55.28 \\
M4&Sentica (+CoS) & 69.89 & 73.25 & 66.06 & 65.91 & 69.67 & 62.35 & 62.24 & 66.66 & 59.40 \\
\rowcolor{lightblue} M5&Sentica (+CoS+PpV)  & \textbf{72.57} & \textbf{76.18} & \textbf{69.39} & \textbf{68.68} & \textbf{72.41} & \textbf{65.46} & \textbf{65.75} & \textbf{69.45} & \textbf{62.52} \\
\bottomrule
\end{tabular}
\vspace{-2mm}
\label{tab:main2}
\end{table*}

\vspace{-1mm}
\section{Experiments}

\subsection{Settings}
\vspace{-2mm}
\paratitle{Evaluations.}
For Task-I, we follow DiaASQ~\cite{li-etal-2023-diaasq}, considering evaluation under three dimensions: 1) element-wise detection; 2) pair-wise extraction; 3) overall sextuple extraction.
For the explicit elements, we use the \emph{exact match} F1 metric.
For the implicit elements, we use the \emph{binary match} F1, where we use GPT-4 to evaluate if the gold element is semantically identical to the prediction (1 if yes, otherwise 0).
Since a correct rationale element may not need to strictly match gold term boundaries (i.e., only coinciding with the critical part), 
we take the \emph{proportional match} F1 for its evaluation.
For the compound evaluation, a pair or overall sextuple is correct only when all elements are correct.
Here, the score for rationale above 0.5 is deemed a correct prediction. 
For sextuple extraction, \emph{micro F1} evaluates the entire sextuple, while \emph{identification F1} measures the sextuple without sentiment polarity.
For subtask-II, we mainly measure three targets: 1) if both Initial Sentiment \& Flipped Sentiment (Flip) are correct, 2) if the flipping trigger's category (Trig) is correct, and 3) if both Flip-Trig is correct simultaneously. For (1) and (3), we use \emph{exact match F1}; for (2), we adopt \emph{macro F1}.

\paratitle{Baselines.}
Since no prior research or methods can be directly adopted here for comparisons, we consider maintaining several baselines via our implementations. 
We first retrofit the UGF~\cite{yan-etal-2021-unified} and DiaASQ~\cite{li-etal-2023-diaasq} so that they can execute the multimodal sextuple extraction tasks, where the small-size LMs are used, e.g., Multilingual BERT (Base)~\cite{devlin2018bert} and mT5 (XXL)~\cite{xue2020mt5}.
We also consider existing MLLMs (supporting T/A/I/V) for comparisons, including Unified-IO 2~\cite{lu2023unified} and NExT-GPT~\cite{wu24next}.
All systems are fine-tuned using the PanoSent training set for fairness.

\paratitle{Implementations.}
Given the varying capabilities of different LLMs across languages, we use Flan-T5 (XXL) for English data, ChatGLM2 6B for Chinese data, and Vicuna 7B for Spanish data.
Our Sentica is tuned via LoRA~\cite{hu2021lora}, allowing for the least parameter updating.
The experiments were conducted on hardware with 8*A100 GPUs, each boasting 80GB of memory. 
To ensure the reliability and reproducibility of our results, we tune the system on a developing set and used five different random seeds, selecting our experimental outcomes based on the average scores from five runs. 

\vspace{-2mm}
\subsection{Main Results}

\vspace{-2mm}
\paratitle{Performances on Panoptic Sentiment Sextuple Extraction.} 
Table~\ref{tab:main1} compares the performances of different methods on Subtask-I, where we can gain the following observations. 
First, due to the presence of many implicit elements in our data, the performance of extraction-based baselines (such as DiaASQ and UGF) can be inferior. 
The generative nature of LLM-based methods, however, effectively addresses this, resulting in overall better performance. 
Comparing the performance of Sentica with Unified-IO 2 and NExT-GPT (M3\&4 vs. M5), we see that our method performs better. 
Sentica, when equipped with the CoS framework, shows significant improvement over the direct prompting paradigm (M7 vs. M5).
Moreover, comparing M6 and M7 shows a clear advantage of our proposed CoS reasoning framework over the vanilla CoT method.

Most importantly, when Sentica combines both the CoS and PpV mechanisms, the complete system (M8) exhibits the strongest global performance. 
As seen, across different task evaluation granularities and languages, our system achieves the best scores. 
In both ZH and SP languages, our system also demonstrates a significant superiority over the Sentica CoT-based variant.
Finally, we can observe task evaluation from different perspectives. 
For different elements, the identification of the holder and target is more accurate, while the determination of rationale is more challenging. 
Similarly, the recognition of sentiment-rationale pairs is also more difficult. 
The overall identification of sextuples poses the greatest challenge, providing a challenging benchmark for follow-up research.

\begin{figure}[!t]
\includegraphics[width=0.98\columnwidth]{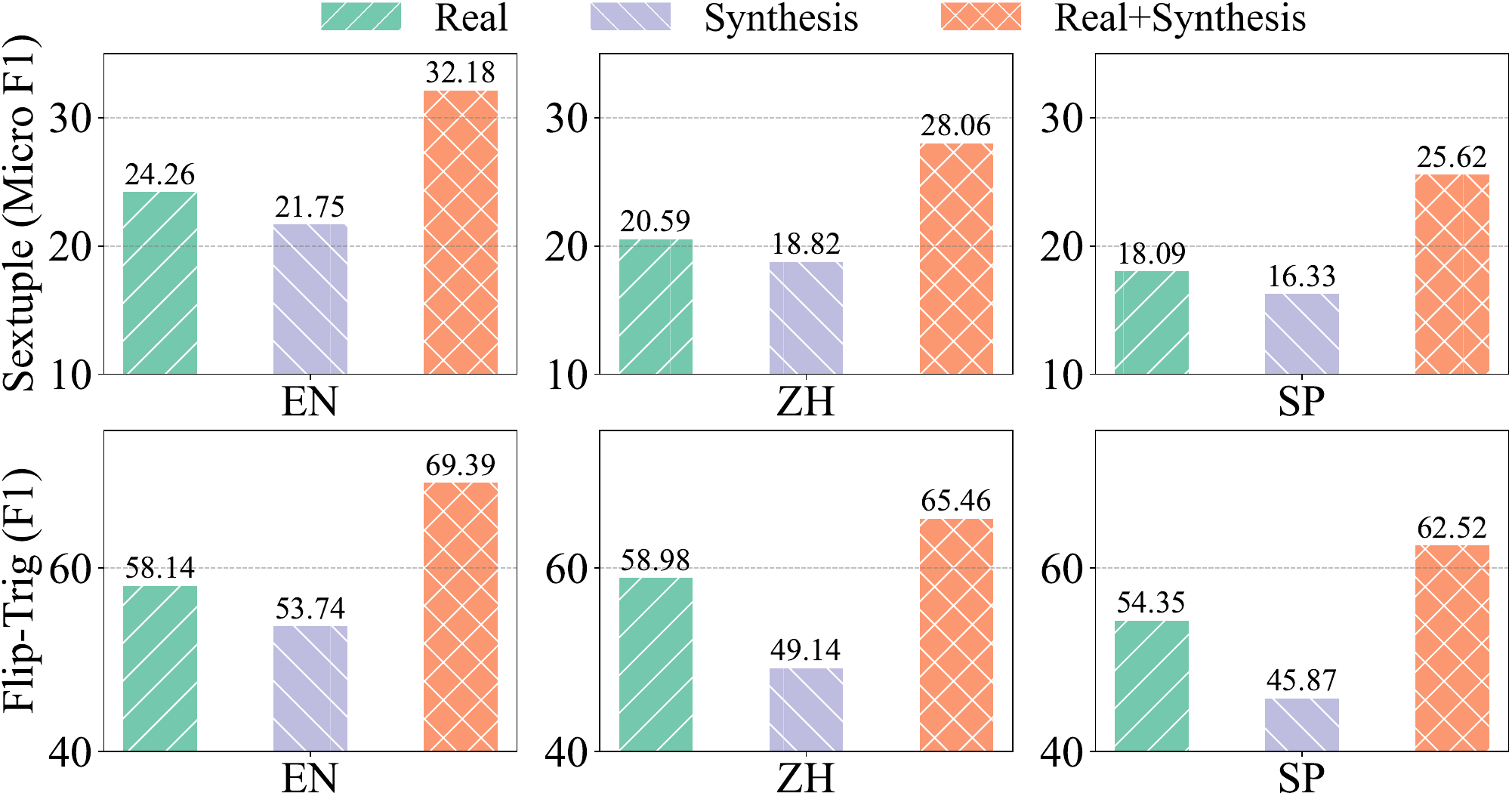}
\vspace{-2mm}
\caption{
Performance with different data sources.
}
\vspace{-4mm}
\label{datasource}
\end{figure}

\begin{figure}[!t]
\includegraphics[width=0.98\columnwidth]{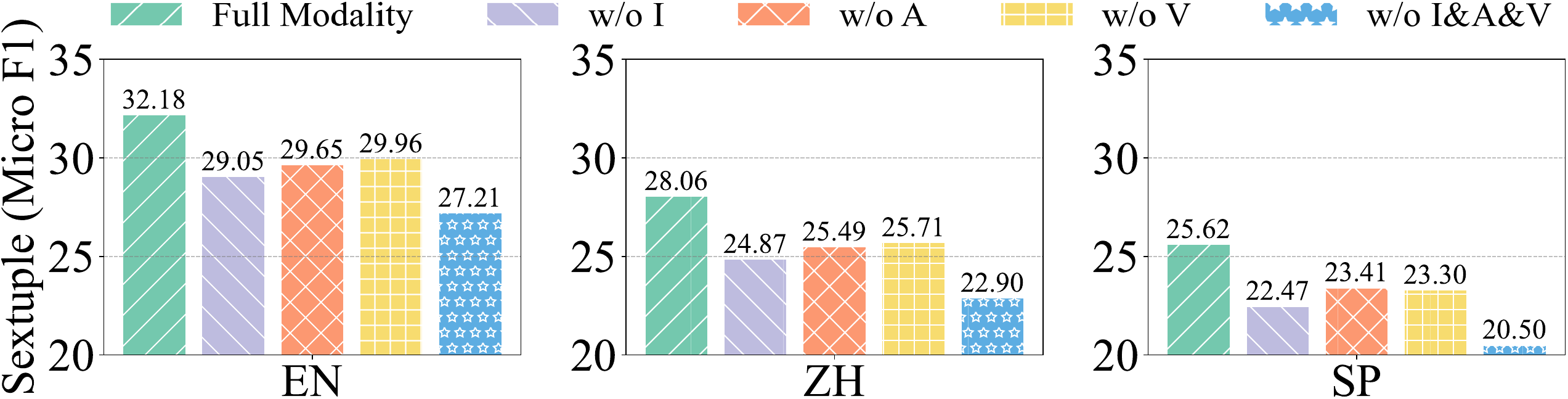}
\vspace{-2mm}
\caption{
Evaluation of the contribution of each modality.
}
\vspace{-4mm}
\label{w/o_modality}
\end{figure}

\paratitle{Results on Sentiment Flipping Analysis.}
For Task 2, we present the overall results in Table~\ref{tab:main2}. 
Similar trends to those observed in Table~\ref{tab:main1} are evident. For instance, our Sentica, on the same backbone LLM, outperforms NExT-GPT. 
Additionally, the CoS reasoning approach, compared to direct prompting or the CoT technique, significantly enhances the accuracy of sentiment flipping identification across all languages. Moreover, our complete system (i.e., Sentica+CoS+PpV) demonstrates the best performance.
The main results and observations from the above two subtasks evidently demonstrate the effectiveness of our proposed methods.

\vspace{-2mm}
\subsection{Analysis and Discussion}

We take one step over the overall performance, further delving into the analyses of the proposed data and methods.

\vspace{-1mm}
\paratitle{Q1: Is It Necessary to Construct Synthetic Data?}
In the above experiments, we train the model by combining real data with synthetic data. 
Therefore, we plan to train the model using these two types of data separately and compare the performance. 
The results for the two subtasks under different languages are shown in Figure~\ref{datasource}. 
Overall, it is observable that training on real-life data yields better results compared to training on synthetic datasets, even though the latter are more plentiful. 
This is because real data possess a more authentic distribution of information, enabling the model to learn a richer set of features. 
Moreover, our test set is also sampled from real data.
Most importantly, we discover that once synthetic data is used as an additional supplement to substantially expand the quantity of real data, it can significantly enhance the final performance, consistently. 
This proves the necessity to construct synthetic data.

\begin{figure}[!t]
\includegraphics[width=0.98\columnwidth]{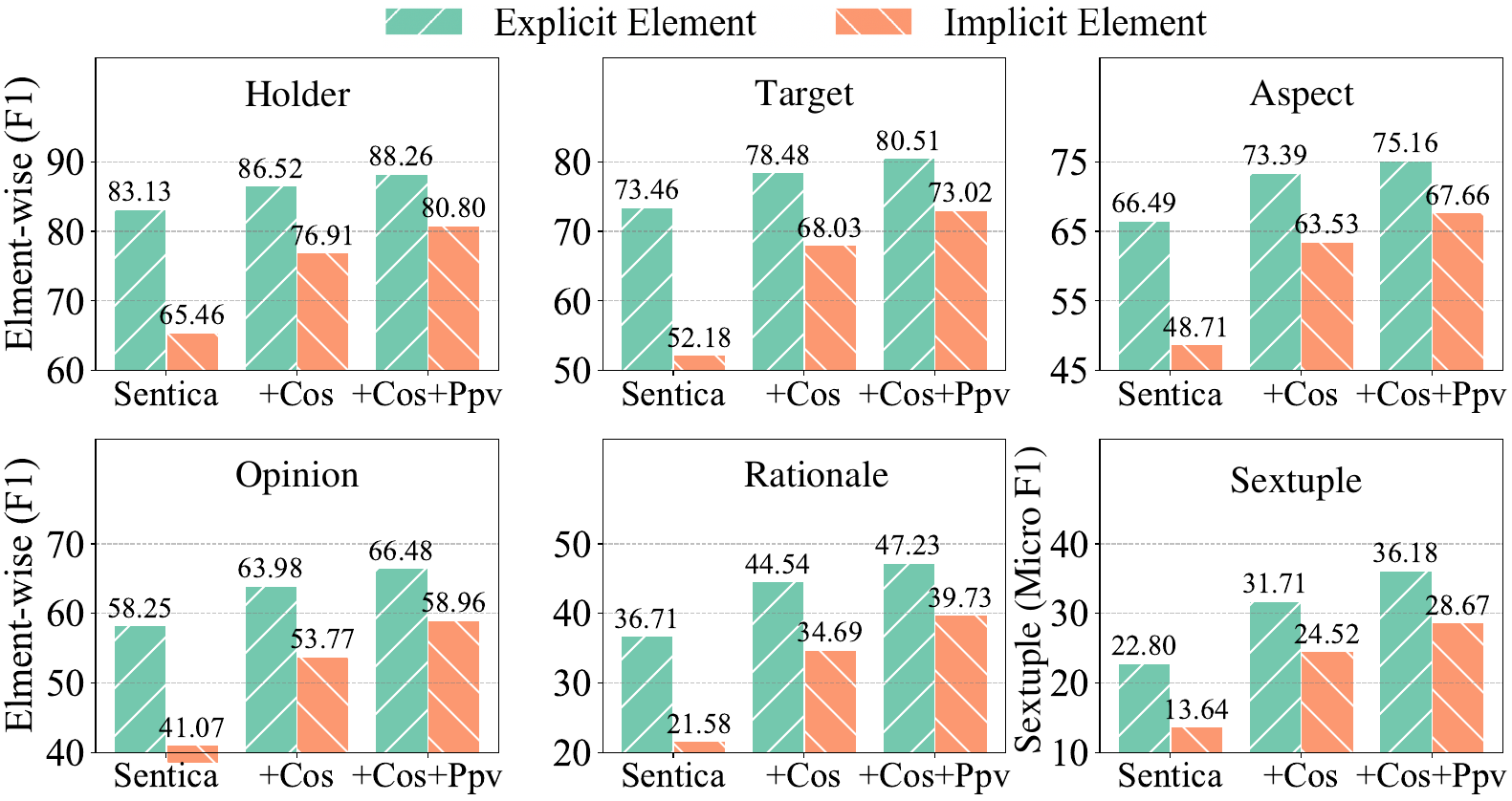}
\vspace{-2mm}
\caption{
Analysis of Explicit and Implicit Elements.
}
\vspace{-2.5mm}
\label{exp_imp}
\end{figure}

\paratitle{Q2: How Significant Is the Role of Multimodal Information?}
Although multimodal information has been utilized in existing multimodal sentiment analysis research~\cite{marrese-taylor-etal-2020-multi,lai2023shared}, it is mostly treated as supplementary to textual information for aiding in the determination of sentiment polarity. 
In this work, the role of multimodal information is comprehensive and all-encompassing. 
It not only assists in determining sentiment polarity but also serves as a direct source of information for judging the sextuple elements (i.e., in an implicit manner).
We demonstrate the impact of removing multimodal information from the test set on the performance of the sextuple extraction task in Figure~\ref{w/o_modality}. 
As seen, removing any type of modal signal results in a downgrade in performance, with the information from images being the most crucial. 
Removing all non-text modalities has the most significant impact.

\paratitle{Q3: How Are Performance for Explicit and Implicit Elements Individually?}
We define the sextuple extraction wherein elements can either be explicitly derived from text or implicitly inferred from contexts or various modalities. 
While the overall results previously presented combine the performance of both explicit and implicit elements, here we aim to showcase the specific performance of various elements individually.
As presented in Figure~\ref{exp_imp}, the performance of implicit elements is consistently lower than that of explicit elements. 
This indicates that recognizing implicit elements is much more challenging. 
This phenomenon aligns with reality; because, compared to extracting explicit text, identifying implicit elements requires a comprehensive understanding of the context's semantics before inferring the corresponding elements.

\paratitle{Q4: Is the PpV Mechanism Reasonable?}
Lastly, we verify the rationality of the proposed PpV mechanism. 
We adopt a template-based approach for paraphrasing $k$-tuples, then check whether the semantics of the structured data coincide with the given context of dialogue. 
In Table~\ref{tab:PpV-eval}, we present some evaluations.
We explore the task performance under different mechanisms, including paraphrasing via LLM, direct verification without paraphrasing, and without any verification. 
It is evident that the PpV mechanism outperforms both direct verification and no verification.
Furthermore, for PpV, we conduct entailment detection between the obtained paraphrases and the dialogue context through human evaluation and then report the accuracy. 
We see that using fixed templates for paraphrasing is more reliable than utilizing LLMs to paraphrase structured tuples.

\begin{table}[!t]
\fontsize{8}{9.5}\selectfont
\setlength{\tabcolsep}{1mm}
\centering
\caption{Comparison of different verification mechanisms.}
\vspace{-3mm}
\begin{tabular}{@{} lccc @{}}
\toprule
& \multicolumn{2}{c}{Task F1}& \multicolumn{1}{c}{Human Acc.}\\
\cmidrule(lr){2-3} \cmidrule(lr){4-4}
& Sextuple & Flip-Trig & Entail Detect.\\
\midrule
\rowcolor{lightblue} PpV (paraphrase via template) & \textbf{32.18} & \textbf{69.39} & \textbf{88.15} \\
PpV (paraphrase via LLM) & 30.83 & 67.60 & 73.62 \\
\hdashline
dir. verify & 30.26 & 67.04 & /  \\
\hdashline
w/o verify & 29.71 & 66.06 & /  \\
\bottomrule
\end{tabular}
\vspace{-5mm}
\label{tab:PpV-eval}
\end{table}

\section{Conclusion}

This paper introduces a novel multimodal conversational ABSA, where the Panoptic Sentiment Sextuple Extraction (including holder, target, aspect, opinion, sentiment, and rationale) and the Sentiment Flipping Analysis tasks are proposed, providing a comprehensive and panoptic definition of sentiment analysis that aligns with the complexity of human-level emotional expression and cognition.
We benchmark the novel settings with PanoSent, a large-scale high-quality dataset annotated both manually and automatically, featuring conversational contexts, multimodality, multilingualism, and multi-scenarios.
We then benchmark the tasks with an effective Chain-of-Sentiment reasoning framework, together with a novel MLLM (namely Sentica) and a paraphrase-based verification mechanism, serving as a strong baseline for subsequent research.

\clearpage

\begin{acks}
This work is supported by the Ministry of Education, Singapore, under its MOE AcRF TIER 3 Grant (MOE-MOET32022-0001).
\end{acks}

\bibliographystyle{ACM-Reference-Format}
\bibliography{sample-base}

\newpage

\appendix

\section{What To Do Next with PanoSent?}

In this paper, we introduce a novel benchmark for Multimodal Conversational Aspect-based Sentiment Analysis, which includes two innovative subordinate tasks: Panoptic Sentiment Sextuple Extraction and Sentiment Flipping Analysis. 
We have proposed the Chain-of-Sentiment reasoning method based on our MLLM, which has demonstrated strong benchmark performance on our dataset, PanoSent. 
We firmly believe that this pioneering work will inaugurate a new era for the sentiment analysis community. 
Several important directions for future research can emerge from our work.

\paratitle{\(\blacktriangleright\) Exploring Multimodality in PanoSent}
In this paper, we encode multimodal information in a straightforward manner using common techniques. Given the critical role of multimodal information for this task, future efforts should focus on developing more powerful methods for multimodal feature extraction and integration. Additionally, investigating the impact of different modalities on sentiment recognition across various scenarios promises to be a fruitful area of research.

\paratitle{\(\blacktriangleright\) Identifying Implicit Sentiment Elements}
Compared to explicit sentiment elements, the identification of implicit elements poses a greater challenge. 
Our approach, based on MLLM, autonomously determines the recognition of implicit sentiment elements through an understanding of the input data's content. 
We believe there are more accurate methods to be discovered for identifying implicit elements.

\paratitle{\(\blacktriangleright\) Sentiment Cognition and Reasoning Mechanisms}
Our new task involves complex sentiment cognition, for which we propose a reasoning framework. 
Future research should delve deeper into the mechanisms of interaction and triggering among sentiment elements, as well as the mechanisms behind Sentiment flipping, in order to develop more robust sentiment reasoning solutions.

\paratitle{\(\blacktriangleright\) Modeling Dialogue Context}
Dialogue scenarios closely resemble the natural ways in which people express emotions. This work processes the overall content of dialogues through the model, allowing it to understand conversations autonomously. 
Next steps in research could focus on how to more effectively enhance the model's ability to model dialogue context, thus better addressing cross-utterance issues. 
For example, further consideration could be given to modeling dialogue structure and speaker coreference resolution features.

\paratitle{\(\blacktriangleright\) Sentiment-aware Instruction Fine-tuning}
Our work involves tasks based on a MLLM, which is fine-tuned on our training set. 
Research indicates that the setup of instruction fine-tuning significantly affects the LLM's performance on downstream tasks. 
We believe that developing superior methods for instruction fine-tuning, such as designing approaches that increase the LLM's sensitivity to sentiment, holds great promise.

\paratitle{\(\blacktriangleright\) Cross-lingual Transfer Learning}
Our dataset includes three popular languages from different language families: English, Chinese, and Spanish, with non-parallel annotations across languages. 
Subsequent research could explore cross-lingual transfer learning in a multimodal scenario, investigating the supportive role of language-invariant features (multimodal information) for sentiment learning across languages.

\paratitle{\(\blacktriangleright\) Cross-domain Transfer Learning}
Our dataset is extensive, covering hundreds of different domains and everyday scenarios. 
It would be interesting to study the variations of panoptic sentiment across different scenes and domains, making cross-domain transfer learning a meaningful direction for future work.

\paratitle{\(\blacktriangleright\) Weak/Unsupervised Sentiment Analysis}
Our paper primarily focused on supervised learning using a large amount of annotated data. However, MLLMs already possess significant unsupervised generalization capabilities. 
It is crucial to leverage our benchmark for weak or even unsupervised sentiment recognition. 
In the subsequent part of the Appendix, we provide an analysis and exploration of few-shot sentiment recognition.

\section{Ethic Considerations}

In conducting this research and developing the \texttt{PanoSent} benchmark, several ethical considerations have been taken into account to ensure the responsible use and application of the technologies involved.

\paratitle{\(\blacktriangleright\) Privacy and Data Protection}
Given that the raw dataset includes multimodal dialogues that may contain personal information, rigorous measures have been implemented to anonymize and protect any potentially sensitive data. 
This includes the removal of personally identifiable information (PII) from texts, images, audio, and video content. Additionally, the dataset has been reviewed to ensure compliance with relevant data protection regulations such as GDPR and CCPA, aiming to respect user privacy fully.
Our data collection procedures have been carefully designed to focus on factual knowledge acquisition without infringing on privacy rights, thereby upholding our strong commitment to privacy and ethical research standards.

\paratitle{\(\blacktriangleright\) Data Collection}
For the creation of the \texttt{PanoSent} dataset, all data was collected from publicly available sources or through contributions from individuals who were informed about the purposes of the research and provided their explicit consent. Efforts were made to ensure that contributors understood their rights, including the right to withdraw their data at any point.

\paratitle{\(\blacktriangleright\) Annotator and Compensation}
Acknowledging the significant role of human annotators in the creation of the \texttt{PanoSent} dataset, we have engaged a diverse group of annotators including well-trained individuals from crowdsourcing platforms, native speakers, and senior postgraduate students with specialized training for the annotation tasks. The estimated time required for annotating each dialogue utterance is between 4 to 6 minutes, reflecting the complexity and detailed nature of the task.

\paratitle{\(\blacktriangleright\) Intellectual Property Protection}
The \texttt{PanoSent} dataset includes content collected from publicly available sources on a popular Chinese social media platform, utilizing its officially open API. This collection method ensures compliance with intellectual property laws and respects the terms of service of the platform. Permission for the use, distribution, and modification of this content is granted under the terms of the Weibo API distribution agreement. This approach safeguards the intellectual property rights of the content creators while facilitating academic research and development.

\paratitle{\(\blacktriangleright\) Bias and Fairness}
Recognizing the potential for bias in AI systems, this research includes an analysis of the \texttt{PanoSent} dataset for biases related to gender, ethnicity, language, and other sociodemographic factors. Steps have been taken to mitigate these biases through diverse and representative data collection across multiple languages and scenarios. However, it is acknowledged that complete eradication of bias is challenging, and continuous efforts are required to identify and address biases as the benchmark evolves.

\paratitle{\(\blacktriangleright\) Misuse Potential}
The research team is aware of the potential misuse of sentiment analysis technologies, such as applications in surveillance or the manipulation of public opinion. Therefore, alongside the release of the \texttt{PanoSent} benchmark and the associated models, guidelines have been developed to encourage ethical use. These guidelines emphasize the importance of consent, transparency, and accountability in any application or further development of the technologies presented in this paper.

\paratitle{\(\blacktriangleright\) Accessibility and Inclusivity}
In line with our commitment to fostering an inclusive research community, all code and data related to the \texttt{PanoSent} benchmark will be made openly available. This ensures that researchers and practitioners from diverse backgrounds and with varying levels of resources have equal opportunities to contribute to, and benefit from, the advancements in multimodal conversational aspect-based sentiment analysis.

\section{More Details of Datasets}

\subsection{Extended Details of Data Construction}
\label{Extended Details of Data Construction}

\paratitle{C.1.1 Data Acquisition}

\begin{figure}[!t]
 \centering
 \includegraphics[width=\columnwidth]{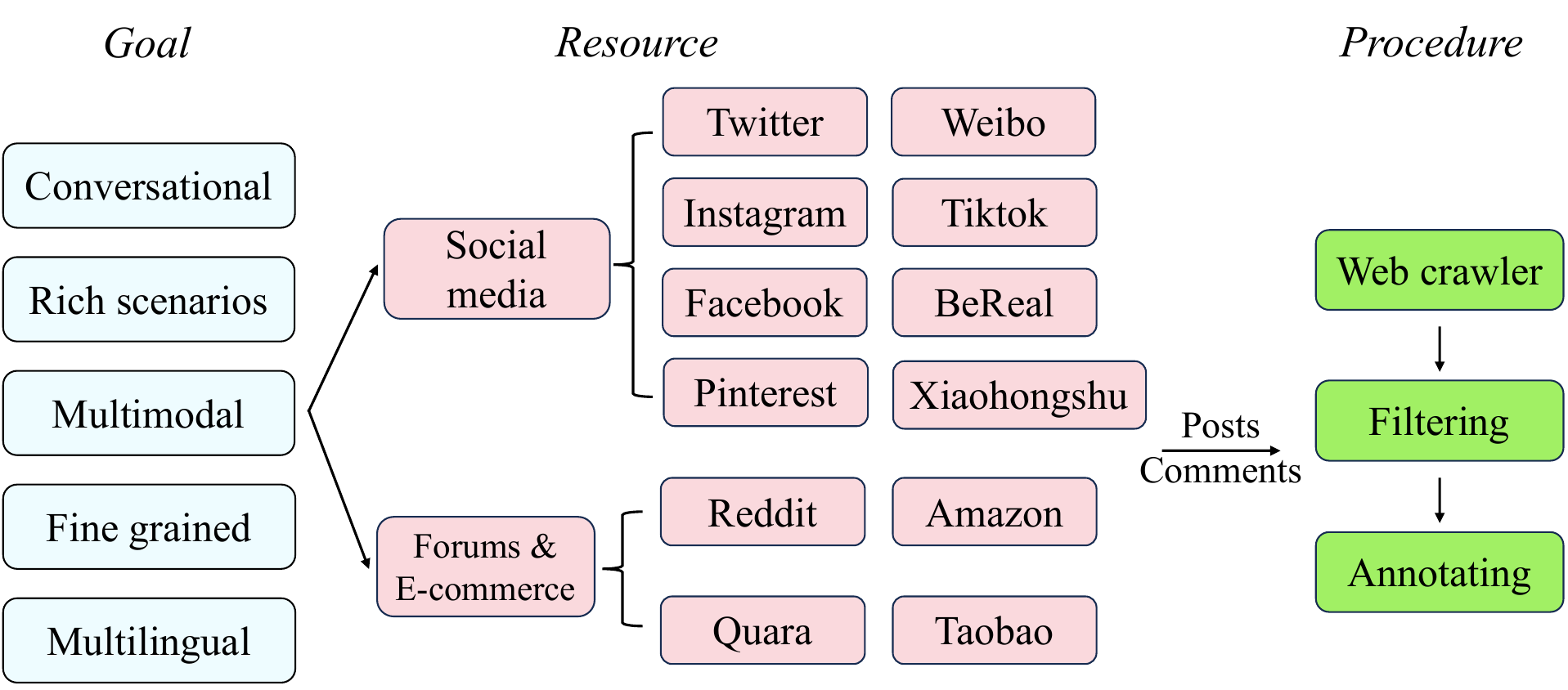}
 \caption{The workflow of data acquisition and preprocessing.}
 \label{datacollect}
 \vspace{-4mm}
\end{figure}

\textbf{$\blacktriangleright$ Step1. Platform Selection and Data Collection.}
Our initial step involves identifying a diverse range of social media and forum platforms as sources for our dataset, including but not limited to Twitter, Facebook, Reddit, Weibo, Xiaohongshu, BeReal. 
These platforms are chosen for their rich conversational content across multiple languages and the vast user engagement they facilitate. 
We target some influential bloggers within specific domains and the discussions surrounding trending topics related to our research themes. Conversations on these platforms typically originate from a root post, with users participating in multi-thread and multi-turn dialogues based on the initial post. 
In addition to text, these interactions often include multimodal content such as images, videos, and audios. 
While less common than text, this multimodal interaction is a crucial component of our dataset, and we make extra efforts to collect conversations incorporating these elements. 
Given that these platforms generally do not support audio replies as a standalone feature, we extract the audio tracks from video content to collect audio modal information. 
Data collection is automated through publicly available APIs provided by these platforms, with conversations being categorized based on their thematic relevance and the types of modal information they contain.
The process of data acquisition and preprocessing is depicted in Figure~\ref{datacollect}.


\textbf{\(\blacktriangleright\) Step2. Data Cleaning and Re-organization.}
To ensure the dataset is free from harmful content, privacy violations, irrelevant, or low-quality conversations, we employ a combination of manual inspection and automated tools. 
A keyword library is constructed based on previous related studies and the expertise of team members in social media analysis and specific thematic areas. 
This library includes keywords indicating potential harm, privacy infringement, and irrelevance to the research topic. 
Scripts are developed to automatically scan the collected data for these keywords, with flagged conversations undergoing manual review to determine their suitability for inclusion in the dataset. 
Additionally, we utilize the Toxic BERT model, capable of identifying various forms of harmful speech, including insults, discrimination, and harassment, by analyzing extensive online textual data. 
This model provides probability scores for detected categories and identifies the specific locations of toxic speech within the text. 
The output from the model is also subject to manual review, considering the context of the conversations to make final decisions on content inclusion. 
Multimodal content is manually reviewed due to its relatively lower volume, focusing not only on the potential harm but also on the relevance of the content to the conversation, with any mismatched multimodal content being removed.

\paratitle{C.1.2 Human Annotation}


We have recruited a team of annotators who possess relevant background knowledge, including well-trained individuals from crowdsourcing platforms, native speakers, and senior postgraduate students. 
Before commencing manual annotation, we developed detailed annotation guidelines based on the definitions from SemEval related to ABSA and the specific requirements of our task. 
All annotators have undergone uniform training to ensure consistency and objectivity in their work. 
Based on the task's complexity and the time needed for careful annotation, we estimate that annotators will require 4 to 6 minutes per data entry. 
Each piece of data has been annotated by at least three independent annotators, and we have calculated the Cohen's Kappa Score to measure the consistency among them. Achieving a score of 0.85, which reflects the high quality of our annotated dataset, data with Kappa scores below a predefined standard undergo review and discussion. 
In cases of disagreement or ambiguity, linguists and native speakers collaborate to reach a consensus. 
Data that cannot reach consensus or remains ambiguous is discarded to maintain the quality of the dataset.

\paratitle{C.1.3 Automatic Synthesis}


Our task mandates rigorous data requirements, necessitating dialogue context that is fine-grained enough to encompass all six defined elements, includes both implicit and explicit expressions, and incorporates multimodal information. Given that only a minuscule proportion of real-world data meets these criteria, and considering the proven success of LLMs in generating data, we have opted to utilize the capabilities of GPT-4 for automated data generation and corresponding element annotation. The process unfolds in several steps.


\textbf{\(\blacktriangleright\) Step1. Creation of Dialogue Instances.}
Drawing from high-quality real dialogues, we meticulously crafted a small batch of dialogue instances tailored to our task's needs. 
These instances display diversity in themes, participant count, length, turn-taking, reply structure, and types of included multimodal information. 
They undergo multiple rounds of modification and inspection by our team to ensure comprehensive coverage and quality.


\textbf{\(\blacktriangleright\) Step2. Prompt Template Design and Data Generation.}
We develop structured and coherent prompt templates to guide GPT-4\footnote{\emph{gpt-4-1106-preview} version API, \url{https://openai.com/gpt-4}} in understanding our requirements and generating dialogue data that aligns with them. After several iterations of adjustments and tests, we finalize a prompt template. 
This template instructs GPT-4 not only to generate dialogues but also to annotate them with the defined sextuples and identify instances of sentiment flips. 
Moreover, for certain dialogue utterances, GPT-4 is tasked with creating suitable captions as placeholders for images, audios, and videos, reflective of the context.

\begin{tcolorbox}[breakable, fontupper=\customfont]
\vspace{-2mm}
{\small
\textbf{An example of our prompt template}: 

As an expert playwright skilled in crafting dialogues, your task is to generate conversations centered around the theme `Televisions'. Please comply with the following instructions. Do not comment, judge, or output other texts and only return the results. \\
1. Generate a nonlinear dialogue replying structure among 4 speakers, and the turns of the dialogue must be 2.\\
2. Each speaker in the dialogue should have a unique `speaker\_id' and a unique `speaker\_name', and each dialogue should have a unique `doc\_id'.\\
3. The dialogue should revolve around one, two, or three main targets (the objects being discussed). For these targets, the conversation should focus on specific aspects (attributes or features of the targets) and provide an opinion (evaluation of the aspect). Each utterance must include an opinion about an aspect and be supported by a rationale (reason or explanation for the opinion). \\
4. Use your creativity and content generation skills to add image modalities in the conversation. The image caption must provide a concrete description of the visual content, detailing the objects, scenes, or actions depicted in the image. The caption must be directly related to the utterance content and should not be vague or abstract. If an image is included, specify 'type' as 'img', 'caption' as the detailed image description, and 'id' as a unique identifier. \\
5. Every utterance except the first utterance is a reply to dialogue sentence with index n, the reply property of this utterance should be n, the first utterance is -1.\\
6. The conversation must include all four elements: `target`, `aspect`, `opinion`, and `rationale`. Annotate and `order` the occurrence of these elements in HTML format in the `annotation`. All elements must be explicitly mentioned in the dialogue text and marked as `explicit`.\\
7. Store all parts of the conversation in accordance with the provided example format. For each utterance, the 'modality' should be set to `None` or include the `type`, `caption`, and `id` if an image is used. \\
8. Ensure full comprehension of the provided example and apply it to create a dialogue that meets all specified criteria, including the proper integration of multimodal elements. Adhere strictly to the example `json` format for organizing the storage structure of the generated dialogue, as shown in the provided example.\\
For instance, a sample `json` output would be: \{sample\_json\_string\} \\}

\vspace{-4mm}
\end{tcolorbox}


\textbf{\(\blacktriangleright\) Step3. Multimodal Information Retrieval.}
With the annotated dialogues, we use the captions to retrieve the piece of information in the corresponding modality (image, audio, or video) from extensive databases such as COCO, Flickr30k for images, AudioSet and WaveText5K for audios, and WebVid for videos. 
These databases, rich in (image, audio, or video)-caption pairs, enable us to match dialogue captions with database captions using SentenceTransformer\footnote{\url{https://huggingface.co/sentence-transformers}}, focusing on the top-10 most similar candidates for each modality. 
For the associated multimodal content, three annotators score each of the ten candidates on a scale of 1-10. 
The content with the highest average score is selected as the definitive multimodal segment. Should none of the candidates meet the desired criteria—indicating a lack of suitable matches within the databases—we resort to direct retrieval from the Google search engine\footnote{\url{https://www.google.com/}} to ensure exhaustive inclusivity.


\textbf{\(\blacktriangleright\) Step4. Manual Review.}
Each generated dialogue, along with annotations related to the two sub-tasks and multimodal content, undergoes a thorough review by at least two staff members. 
Any potentially problematic instances are discarded. 
We also calculate the Cohen's Kappa Score, achieving a score of 0.82, which attests to the consistency and validity of our annotation process.

In Table~\ref{instance}, we illustrate a complete data instance (a conversation) with our annotation (English version is shown).

\begin{table*}[!htbp] 
\fontsize{8}{10.5}\selectfont
\setlength{\tabcolsep}{1.9 mm}
\centering
\caption{A snippet of an annotated data instance in PanoSent dataset.}
\vspace{-3mm}
\begin{tabular}{m{0.1\textwidth}m{0.8\textwidth}}
\hline
\textbf{Key} & \textbf{Value} \\ \hline
Dialogue-ID & 00024 \\ \hline
Dialogue & 1. Ava: I recently purchased a new digital camera, and its image quality is stunning, capturing every detail with such clarity and vibrant colors that photos almost look lifelike. \newline 2. Liam: That sounds amazing! What about its low-light performance? Does it capture sharp and clear images in low-light conditions? \newline 3. Ava: The low-light performance is quite impressive. It captures sharp and clear images even in dimly lit environments. \newline 4. Noah: What about its battery life? \newline 5. Ava: The battery life is disappointing. It drains quickly and requires frequent recharging. \newline 6. Liam: It's worth noting that the camera's advanced features naturally demand more power, which is common for high-performance devices. Compared to similar models, our camera holds up well in terms of battery life, making it a fair trade-off for its quality. \newline 7. Ava: That's a good point. Considering the advanced features and comparing it with other cameras, the battery life does seem acceptable. I hadn't looked at it that way before. \\ \hline
Replies & -1, 0, 1, 0, 3, 4, 5 \\ \hline
Speakers & 0, 1, 0, 2, 0, 1, 0 \\ \hline
Holders & Ava, Liam \\ \hline
Targets & digital camera \\ \hline
Aspects & image quality, low-light performance, battery life \\ \hline
Opinions & stunning, quite impressive, disappointing, holds up well, acceptable \\ \hline
Sextuples & (Ava, digital camera, image quality, stunning, positive, capturing every detail with such clarity and vibrant colors that photos almost look lifelike) \newline (Ava, digital camera, low-light performance, quite impressive, positive, it captures sharp and clear images even in dimly lit environments) \newline (Ava, digital camera, battery life, disappointing, negative, it drains quickly and requires frequent recharging) \newline (Liam, digital camera, battery life, holds up well, positive, compared to similar models) \newline (Ava, digital camera, battery life, acceptable, neutral, considering the advanced features and comparing it with other cameras)\\ \hline
Sentiment Flip & Holder-Target-Aspect: (Ava, digital camera, battery life) \newline Initial Sentiment-Flipped Sentiment: (negative, neutral) \newline Trigger Type: Participant Feedback and Interaction \\ \hline
\end{tabular}
\label{instance}
\end{table*}

\subsection{Detailed Summary of Dataset Insights}
\label{Detailed Summary of Dataset Insights}

Here, we extend the content of Section $\S$\ref{Data Insights} from the main article, to provide a more comprehensive introduction to all the highlights of our dataset.

\paratitle{\(\blacktriangleright\) Panoptic Fine-grained Sentiment Definition.}
Compared to existing ABSA datasets, the PanoSent dataset stands out for its fine-grained and exhaustive annotation of sentiment elements, featuring six key items essential for ABSA: holder, target, aspect, opinion, sentiment, and rationale. 
The `holder' represents the entity expressing the viewpoint, which, despite frequently being the speaker in conversational contexts, can also encompass instances where the holder is not the speaker. 
The `target' pertains to the subject of discussion, such as a digital gadget, a service, or an activity. 
`Aspect' refers to specific attributes or facets of the target, for example, the battery, screen, or camera quality of a smartphone. 
`Opinion' denotes the expressed view or judgement, while `sentiment' captures the emotional polarity associated with the opinion, classified as positive, neutral, or negative. 
Finally, `rationale' elucidates the underlying reasons or justifications that give rise to a particular opinion. 
This meticulous approach to sentiment analysis not only enhances the depth of understanding around each conversational element but also significantly advances the precision and applicability of ABSA methodologies in dissecting and interpreting complex dialogues.

\paratitle{\(\blacktriangleright\) Cognitive Causal Rationale.}
We not only prioritize the identification of sentiment states and the granularity of emotional details within dialogues but also emphasize the significance of understanding the underlying reasons behind expressed opinions. 
Building on this premise, we introduce the rationale element into ABSA for the first time, refining its definition to include a focus on the motivations behind sentiments. 
This approach aids in a more comprehensive analysis from a logical perspective, unveiling the catalysts behind viewpoints and attitudes, thereby enriching the extraction of deeper semantic insights.

\paratitle{\(\blacktriangleright\) Dynamic Sentiment Flipping.}
In the complex scene of dialogues, analyzing dynamic sentiment changes is crucial. 
Participants in a conversation may alter their previous viewpoints and attitudes due to various triggers, a vital aspect for understanding the progression of events and emotional trends within dialogues, such as changes in characters' psychological states. 
This dynamic aspect of sentiment, however, has not been addressed in existing ABSA research. 
To comprehend the intricate dynamics of sentiment within multiparty dialogues, we categorize four distinct and clearly defined types of triggers that can lead to sentiment flips: \textbf{introduction of new information}, \textbf{logical argumentation}, \textbf{participant Feedback and interaction}, and \textbf{personal experiences and self-reflection}. 
Each of these triggers plays a critical role in the natural evolution of sentiment within conversations, providing a deeper insight into the fluid nature of human emotions and thoughts in dialogue contexts.

1) \textbf{Introduction of New Information} encapsulates instances where new data, research findings, news reports, or previously undiscussed information are introduced into the dialogue. Such information can alter or influence participants' understanding or emotional stance toward a topic. 

2) \textbf{Logical Argumentation} involves constructing arguments through logical reasoning and analysis using known information or consensus. This trigger uses structured and persuasive logic to convince participants to adopt a viewpoint through rational analysis.

3) \textbf{Participant Feedback and Interaction} focuses on the direct feedback and interactions among participants in the dialogue, including opposition, questioning, or other forms of direct response. This category emphasizes how direct interpersonal communication can influence shifts in emotional stances.

4) \textbf{Personal Experiences and Self-reflection} covers instances where individuals trigger a change in their emotional stance by describing their own experiences, reflecting on their perceptions or experiences. This trigger is internal, based on personal memories and their current evaluation.

\begin{table*}[!t]
\fontsize{8}{10}\selectfont
\setlength{\tabcolsep}{1.5mm} 
\centering
\caption{Detailed categorization of domains in PanoSent dataset.
}
\label{tab:domain-categorization}
\begin{tabular}{m{0.19\linewidth}|m{0.68\linewidth}}
\hline
\textbf{Principal Domains} & \textbf{Sub-Domains} \\
\hline
Electronic Products & Smartphones, Personal Computers, Televisions, Wearable Technology, Cameras, Audio Systems, Gaming Hardware, Home Automation, Tablets, Drones, Smart Home Devices, E-Readers \\
\hline
Technology & Artificial Intelligence, Blockchain, Virtual Reality, Cybersecurity Measures, Cloud Solutions, Quantum Devices, Robotics, Network Innovations, Sustainable Energy, Advanced Biotech, Space Exploration Technologies \\
\hline
Fashion & High Fashion, Urban Streetwear, Designer Brands, Vintage Apparel, Accessories, Children's Wear, Sportswear, Sustainable and Ethical Fashion, Techwear, Seasonal Collections \\
\hline
Food and Cuisine & Plant-based Cuisine, Global Street Eats, Gourmet Dining, Mobile Food Services, Regional Delicacies, Sweets and Confectionery, Health-conscious Foods, International Fusion, Culinary Skills, Beverage Crafting \\
\hline
Movies and Entertainment & Major Studio Releases, Indie Films, Documentaries, Streaming Originals, Celebrity Culture, Awards Season, Reality Shows, Animation, Genre Cinema, Film Festival, Web Series, Fan Culture and Fandom \\
\hline
Health and Wellness & Mental Health Awareness, Fitness Regimens, Dietary Plans, Mindfulness and Meditation, Retreats for Wellbeing, Holistic Medicine, Beauty and Dermatology, Sleep Science, Nutritional Supplements, Wellness Gadgets \\
\hline
Finance and Economy & Equities Market, Savings and Budgeting, Property Market, Pensions and Retirement, Fiscal Policies, Insurance Schemes, Trading Strategies, Financial Tech, International Commerce, Crypto Assets \\
\hline
Sports and Athletics & Team Sports, Basketball, Racquet Sports, Olympic Disciplines, Adventure Sports, Digital Gaming Competitions, Gymnastics, Aquatic Activities, Motorsport, Outdoor Challenges, E-Sports Technology, Urban Sports and Street Games \\
\hline
Travel and Tourism & Offbeat Adventures, Cultural Expeditions, Green Travel, Opulent Journeys, Economical Excursions, Sea Cruises, Solo Explorations, Family Getaways, Heritage Sites, Gastronomic Tours \\
\hline
Art and Culture & Modern Art, Musical Variations, Performing Arts, Literary Works, Exhibition Spaces, Cultural Celebrations, Photographic Arts, Sculptural and Installations, Traditional Crafts, New Media Art \\
\hline
\end{tabular}
\label{tab:domains}
\end{table*}

\begin{figure}[!t]
 \centering
 \includegraphics[width=\columnwidth]{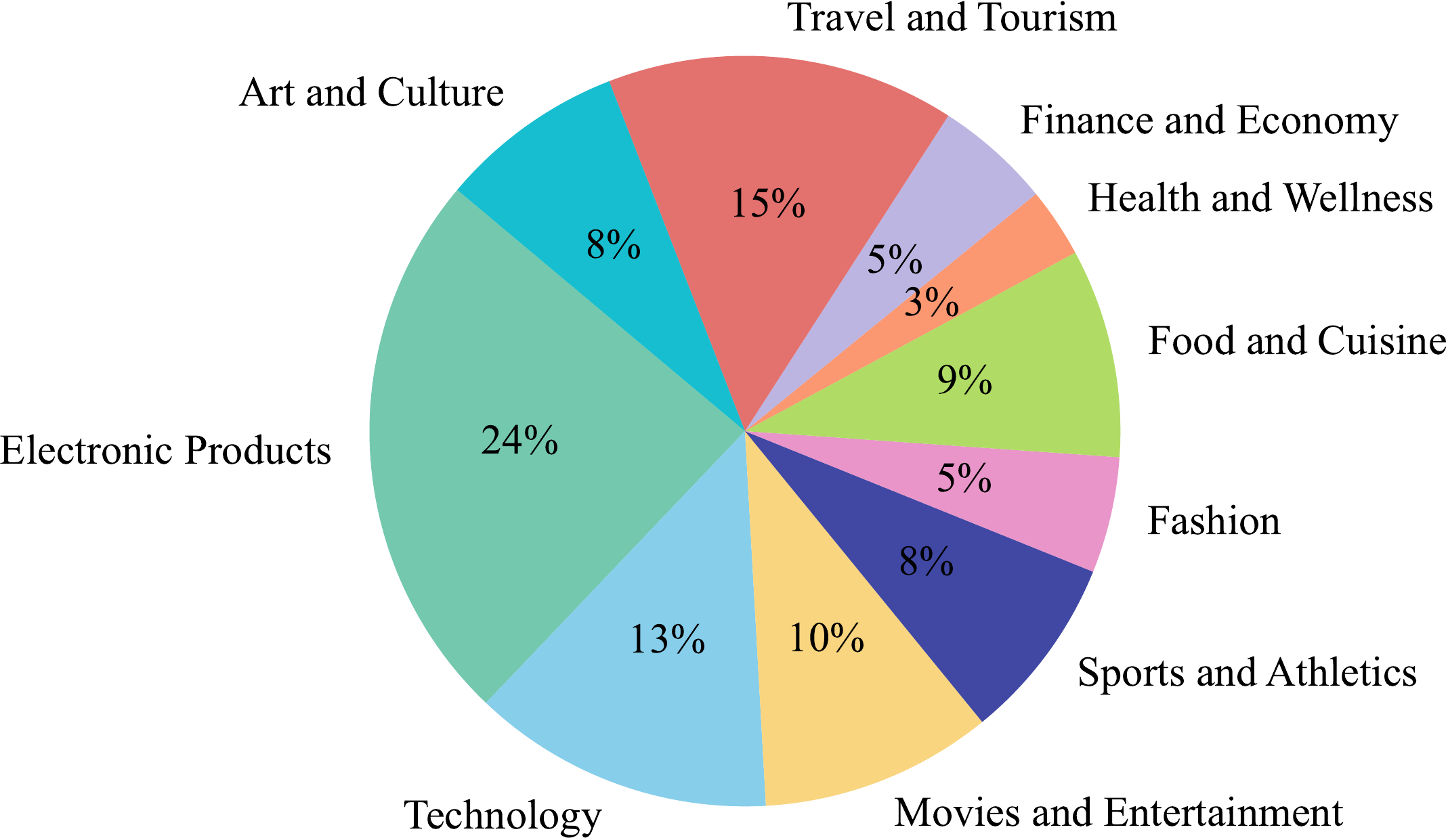}
 \caption{Distribution of categories within each domain.}
 \label{domainprop}
 \vspace{-1mm}
\end{figure}

\paratitle{\(\blacktriangleright\) Multi-scenario.}
PanoSent positions dialogue as its contextual backbone, incorporating 10 primary real-life domains that span over 100 sub-domains, thereby ensuring a broad diversity to facilitate research into sentiment analysis from a variety of perspectives. 
The 10 main domains include electronic products, technology, fashion, food and cuisine, movies and entertainment, health and wellness, finance and economy, sports and athletics, travel and tourism, and art and culture. 
Data within each main domain vary in distribution, and each domain encompasses at least 10 sub-domains. 
The specific classifications and details of these sub-domains are illustrated in Table~\ref{tab:domain-categorization}, while the distribution of categories within each domain is depicted in Figure~\ref{domainprop}.

\paratitle{\(\blacktriangleright\) Multimodality.} 
Our PanoSent dataset showcases a structured amalgamation of multimodal content within dialogues, reflecting the diverse interaction types prevalent in human communication. 
The majority of the dialogues remain text-based. Beyond text, certain dialogues are enriched with images, audios, or videos, thereby integrating visual and auditory dimensions into the textual conversations. The additional modalities include images (22\%), audio (6\%), video (4\%), and mixed modalities (12\%). The mixed modalities encompass combinations like image-audio (IA), image-video (IV), audio-video (AV), and image-audio-video (IAV). 
We ensure these non-textual modalities are abundant, relevant, and of high quality, aligning closely with the dialogue content.

\paratitle{\(\blacktriangleright\) Multilingualism.}
PanoSent encompasses dialogues in three predominant languages: English (60\%), Chinese (30\%), and Spanish (10\%), facilitating cross-lingual research in ABSA. 
To ensure the accuracy and standardization of annotations across each language, we employ online grammar checking tools for preliminary validation of the annotations. Additionally, we engage several native speakers for each language to conduct manual reviews and corrections, guaranteeing that the data annotations are not only standardized but also precise. This meticulous approach ensures the dataset's reliability for cross-lingual sentiment analysis studies.

\paratitle{\(\blacktriangleright\) Implicit ABSA.}
Our dataset comprehensively accommodates implicit ABSA, thereby introducing heightened challenges into the field. Although most sextuple elements are explicitly mentioned in the utterance text, about 28\% of dialogues include implicit elements that need to be inferred from the context of information presented across various modalities.

Contrastingly, most existing studies have predominantly focused on the extraction of explicit elements, largely overlooking the implicit dimensions. 
In reality, whether it's product reviews, daily conversations, or dialogues in other scenarios, a substantial portion comprises implicit elements. 
Hence, implicit elements are exceedingly common and should not be disregarded. 
This emphasis underscores the necessity of integrating both explicit and implicit element analysis to fully capture the nuances and complexities of sentiment in diverse communicative contexts.

\paratitle{\(\blacktriangleright\) Cross-utterance and inner utterance.}
Given that elements of the same sextuple can originate from multiple distinct utterances, potentially spanning across two, three, or even more utterances, the extraction of information spanning multiple utterances poses greater demands on the model's capabilities. 
Our dialogue dataset includes such instances, laying a foundation for subsequent exploration and research. This consideration highlights the intricate dynamics of conversation analysis, emphasizing the necessity for models to adeptly navigate and interpret cross-utterance and inner-utterance relationships to fully understand the context and sentiments expressed.

\begin{figure}[!t]
 \centering
 \includegraphics[width=\columnwidth]{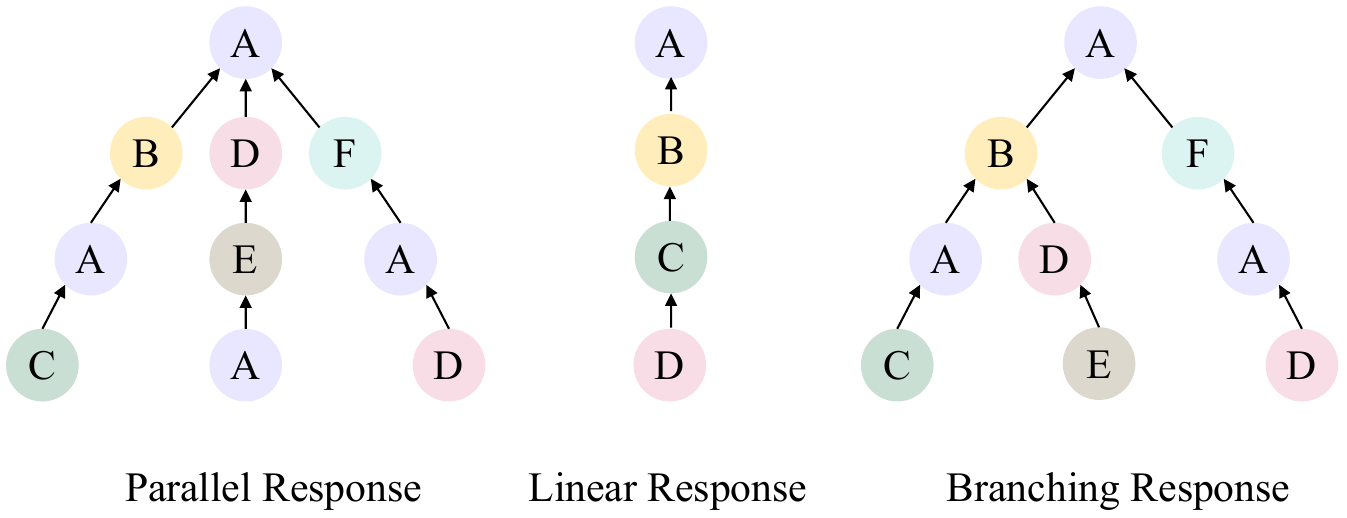}
 \caption{Different replying structure of dialogue.}
 \label{replying}
 \vspace{-1mm}
\end{figure}

\paratitle{\(\blacktriangleright\) Rich dialogue replying structure.}
Commonly, every dialogue starts with a root post, with multiple users (speakers) participating by replying to previous utterances. Consequently, the diversity of a dialogue is manifested not only in superficial distinctions, such as the number of participants or the number of turns within the dialogue but also in the deeper variations of the reply structure. 
We have taken into account the diversity of reply structures and identified three distinct types of reply structures, as illustrated in Figure~\ref{replying}. 
These structures have been carefully considered during the automatic synthesis of dialogues to ensure a realistic and varied representation of conversational dynamics.

\paratitle{\(\blacktriangleright\) High-quality and Large-scale.}
Through meticulous manual annotation and cross-validation, we ensure the high quality of PanoSent.
By employing automated synthesis, we significantly expand the dataset's scale without compromising its quality. This results in a total of 10,000 dialogue instances and 47,100 sextuples.
This high-quality large-volume dataset facilitates subsequent research.




\section{More Details of Methods}


Here, we provide a more detailed introduction to the Chain-of-Sentiment (CoS) reasoning framework and the paraphrase-based verification (PpV) mechanism we proposed. We mainly present more details about the prompts we used.

\subsection{Prompts for CoS Reasoning}
\label{Prompts for CoS Reasoning}


To more clearly illustrate the workflow of our designed CoS mechanism, we provide a specific dialogue example to demonstrate the reasoning process.
The content of the dialogue is as follows:

\vspace{3mm}
\noindent\textit{$\bullet$ [0] Chris: I find the low-light performance is exceptional, capturing clear and vibrant photos even in dim settings. (reply = -1)}

\noindent\textit{[IMAGE$_1$](caption: Dusk light in the forest through a mobile phone lens.)}

\noindent\textit{$\bullet$ [1] Emma: But the battery life to be quite disappointing. It tends to drain quickly even with minimal usage. (reply = 0)}

\noindent\textit{$\bullet$ [2] Sophia: Yes, it is a significant issue, often needing recharging multiple times a day. (reply = 1)}

\noindent\textit{$\bullet$ [3] Lucas: And the phone's design blends elegance with practicality. (reply = 0)}

\noindent\textit{$\bullet$ [4] Chris: However, I don't see it that way; it seems to follow the same formula as its predecessors. (reply = 3)}

\noindent\textit{$\bullet$ [5] Sophia: Have you guys noticed the new model's edge-to-edge display design? It's useful and maximizes screen size without increasing the phone's overall dimensions.(reply = 4)}

\noindent\textit{[VIDEO$_1$](caption: Showcasing the phone's special edge-to-edge display design.)}

\noindent\textit{$\bullet$ [6] Chris: That's a good point. I hadn't really considered that aspect. The edge-to-edge display design is impressive. (reply = 5)}

\vspace{3mm}
Then, the reasoning process of our CoS goes as follows:

\paratitle{\(\blacktriangleright\) Step 1: Target-Aspect Identification.}
\begin{tcolorbox}[breakable, fontupper=\customfont]
\vspace{-2mm}
{\small
\textbf{Input Data}: 

1. Chris: I find the low-light performance is exceptional, capturing clear and vibrant photos even in dim settings. (reply = -1)

2. Emma: But the battery life to be quite disappointing. It tends to drain quickly even with minimal usage. (reply = 0)

3. Sophia: Yes, it is a significant issue, often needing recharging multiple times a day. (reply = 1)

4. Lucas: And the phone's design blends elegance with practicality. (reply = 0)

5. Chris: However, I don't see it that way; it seems to follow the same formula as its predecessors. (reply = 3) 

6. Sophia: Have you guys noticed the new model's edge-to-edge display design? It's useful and maximizes screen size without increasing the phone's overall dimensions. (reply = 4)

7. Chris: That's a good point. I hadn't really considered that aspect. The edge-to-edge display design is impressive. (reply = 5)\\

With encoded information of [IMAGE$_1$], [VIDEO$_1$] \\

\textbf{Instruction}: Based on the multi-party dialogue and its accompanying multimodal data, please identify all possible targets and their specific aspects mentioned in the dialogue. 
Extract each target and aspect explicitly from the utterance text spans, or infer them implicitly via your understanding of the input data. 
Ensure each identified target is paired with its aspect(s), forming target-aspect pairs.\\

\textbf{\color{blue}{Output}}: Target-aspect pairs: (phone, low-light performance), (phone, battery life), (phone, design)} \\

\vspace{-4mm}
\end{tcolorbox}

\paratitle{\(\blacktriangleright\) Step 2: Holder-Opinion Detection.}
\begin{tcolorbox}[breakable, fontupper=\customfont]
\vspace{-2mm}
{\small
\textbf{Input Data}: 

1. Chris: I find the low-light performance is exceptional, capturing clear and vibrant photos even in dim settings. (reply = -1)

2. Emma: But the battery life to be quite disappointing. It tends to drain quickly even with minimal usage. (reply = 0)

3. Sophia: Yes, it is a significant issue, often needing recharging multiple times a day. (reply = 1)

4. Lucas: And the phone's design blends elegance with practicality. (reply = 0)

5. Chris: However, I don't see it that way; it seems to follow the same formula as its predecessors. (reply = 3) 

6. Sophia: Have you guys noticed the new model's edge-to-edge display design? It's useful and maximizes screen size without increasing the phone's overall dimensions. (reply = 4)

7. Chris: That's a good point. I hadn't really considered that aspect. The edge-to-edge display design is impressive. (reply = 5)\\

Target-aspect pairs: (phone, low-light performance), (phone, battery life), (phone, design)\\

\textbf{Instruction}: Based on the dialogue and each target-aspect pair identified previously, please identify the holder (the person who expresses an opinion, normally should be a speaker of certain dialogue utterance) and the opinion, both either directly extracted from the text or inferred from our understanding of the input data.
Formulate your output into `holder-target-aspect-opinion' quadruples, ensuring each element is clearly identified.\\

\textbf{\color{blue}{Output}}: Target-aspect pairs: (Chris, phone, low-light performance, exceptional), (Emma, phone, battery life, quite disappointing), (Sophia, phone, battery life, a significant issue), (Lucas, phone, design, good), (Chris, phone, design, ordinary), (Sophia, phone, design, useful), (Chris, phone, design, impressive)\\
}
\vspace{-4mm}
\end{tcolorbox}

\paratitle{\(\blacktriangleright\) Step 3: Sentiment-Rationale Mining.}
\begin{tcolorbox}[breakable, fontupper=\customfont]
\vspace{-2mm}
{\small
\textbf{Input Data}:

1. Chris: I find the low-light performance is exceptional, capturing clear and vibrant photos even in dim settings. (reply = -1)

2. Emma: But the battery life to be quite disappointing. It tends to drain quickly even with minimal usage. (reply = 0)

3. Sophia: Yes, it is a significant issue, often needing recharging multiple times a day. (reply = 1)

4. Lucas: And the phone's design blends elegance with practicality. (reply = 0)

5. Chris: However, I don't see it that way; it seems to follow the same formula as its predecessors. (reply = 3) 

6. Sophia: Have you guys noticed the new model's edge-to-edge display design? It's useful and maximizes screen size without increasing the phone's overall dimensions. (reply = 4)

7. Chris: That's a good point. I hadn't really considered that aspect. The edge-to-edge display design is impressive. (reply = 5) \\

Holder-target-aspect-opinion quadruples:
(Chris, phone, low-light performance, exceptional), (Emma, phone, battery life, quite disappointing), (Sophia, phone, battery life, a significant issue), (Lucas, phone, design, good), (Chris, phone, design, ordinary), (Sophia, phone, design, useful), (Chris, phone, design, impressive) \\

\textbf{Instruction}: Based on the dialogue and each holder-target-aspect-opinion quadruple identified previously, please identify the sentiment polarity associated with the opinion and analyze the causal rationale behind it. 
The sentiment polarity should be classified as `positive', `neutral', or `negative'. The rationale should be extracted explicitly from the text, or
inferred implicitly via your understanding of the input data.
Formulate your output into `holder-target-aspect-opinion-sentiment-rationale' sextuples, ensuring sentiment polarity is clearly analyzed and the other five elements is clearly identified.\\

\textbf{\color{blue}{Output}}: Holder-target-aspect-opinion-sentiment-rationale sextuples: (Chris, phone, low-light performance, exceptional, positive, capturing clear and vibrant photos even in dim settings), (Emma, phone, battery life, quite disappointing, negative, drain quickly even with minimal usage), (Sophia, phone, battery life, a significant issue, negative, often needing recharging multiple times a day), (Lucas, phone, design, good, positive, blends elegance with practicality), (Chris, phone, design, ordinary, neutral, follow the same formula as its predecessors), (Sophia, phone, design, useful, positive, maximizes screen size without increasing the phone's overall dimensions), (Chris, phone, design, impressive, positive, I hadn't really considered that aspect.)\\
}
\vspace{-4mm}
\end{tcolorbox}

\paratitle{\(\blacktriangleright\) Step 4: Sentiment Flipping Trigger Classification.}
\begin{tcolorbox}[breakable, fontupper=\customfont]
\vspace{-2mm}
{\small
\textbf{Input Data}: 

1. Chris: I find the low-light performance is exceptional, capturing clear and vibrant photos even in dim settings. (reply = -1)

2. Emma: But the battery life to be quite disappointing. It tends to drain quickly even with minimal usage. (reply = 0)

3. Sophia: Yes, it is a significant issue, often needing recharging multiple times a day. (reply = 1)

4. Lucas: And the phone's design blends elegance with practicality. (reply = 0)

5. Chris: However, I don't see it that way; it seems to follow the same formula as its predecessors. (reply = 3) 

6. Sophia: Have you guys noticed the new model's edge-to-edge display design? It's useful and maximizes screen size without increasing the phone's overall dimensions. (reply = 4) 

7. Chris: That's a good point. I hadn't really considered that aspect. The edge-to-edge display design is impressive. (reply = 5) \\

Holder-target-aspect-opinion-sentiment-rationale sextuples: (Chris, phone, low-light performance, exceptional, positive, capturing clear and vibrant photos even in dim settings), (Emma, phone, battery life, quite disappointing, negative, drain quickly even with minimal usage), (Sophia, phone, battery life, a significant issue, negative, often needing recharging multiple times a day), (Lucas, phone, design, good, positive, blends elegance with practicality), (Chris, phone, design, ordinary, neutral, follow the same formula as its predecessors), (Sophia, phone, design, useful, positive, maximizes screen size without increasing the phone's overall dimensions), (Chris, phone, design, impressive, positive, I hadn't really considered that aspect.) \\

\textbf{Instruction}: Based on the dialogue and each holder-target-aspect-opinion-sentiment-rationale' sextuple, please identify instances where a sentiment flip occurs for the same holder regarding the specific target-aspect pair. 
Determine the trigger type for these flips from the predefined categories: \textit{introduction of new information}, \textit{logical argumentation}, \textit{participant feedback and interaction}, \textit{personal experience and self-reflection}.
Formulate your output to include the holder, target, aspect, initial sentiment, flipped sentiment, and the trigger type, or state "None" if no flips are identified.\\

\textbf{\color{blue}{Output}}: (Chris, phone, design, neutral, positive, Introduction of New Information) \\
}
\vspace{-4mm}
\end{tcolorbox}

\subsection{Prompts for Paraphrase-based Verification}
\label{Prompts for Paraphrase-based Verification}

In our paraphrase-based verification mechanism, the transformation of k-tuples into natural language expressions is carefully designed for each specific $k$-tuple. 
This ensures that the expressions accurately reflect the intended sentiment analysis's meaning and context. 
Each step in the verification process can yield multiple outcomes—such as pairs, quadruples, or sextuples—depending on the specific demands of the analysis task.
For example, in the initial step, if $k$ target-aspect pairs are identified, they are represented as $(t_1, a_1), (t_2, a_2), ..., (t_k, a_k)$. 
The verification templates that follow are structured to assess the consistency of these outcomes with the dialogue content, thereby validating the precision of our analysis.

\paratitle{\(\blacktriangleright\) Step 1: Verification of Target-Aspect Identification}
\begin{tcolorbox}[breakable, fontupper=\customfont]
{\small 
\textbf{Input Data}: \(D\), \(\{(t_i, a_i)\}\) \\
\textbf{Instruction}: In this dialogue, participants discussed various targets and their corresponding aspects, including $a_1$ of $t_1$, $a_2$ of $t_2$, etc. 
Please based on the dialogue, verify whether these descriptions are consistent with the dialogue content and provide `1' for `yes' or `0' for `no' judgment.\\

\textbf{\color{blue}{Expected Output}}: 1 (if yes) or 0 (if no)}
\end{tcolorbox}

\paratitle{\(\blacktriangleright\) Step 2: Verification of Holder-Opinion}
\begin{tcolorbox}[breakable, fontupper=\customfont]
{\small
\textbf{Input Data}: \(D\), \(\{(h_j, t_i, a_i, o_j)\}\) \\
\textbf{Instruction}:In this dialogue, different participants expressed their opinions towards various aspects of targets, including the opinion of $h_1$ on $a_1$ of $t_1$ is $o_1$, and the opinion of $h_2$ on $a_2$ of $t_2$ is $o_2$, etc. 
Please based on the dialogue, verify whether these descriptions are consistent with the dialogue content and provide `1' for `yes' or `0' for `no' judgment.\\

\textbf{\color{blue}{Expected Output}}: 1 (if yes) or 0 (if no)}
\end{tcolorbox}

\paratitle{\(\blacktriangleright\) Step 3: Verification of Sentiment-Rationale Mining}
\begin{tcolorbox}[breakable, fontupper=\customfont]
{\small
\textbf{Input Data}: \(D\), \(\{(h_j, t_i, a_i, o_j, s_k, r_l)\}\) \\
\textbf{Instruction}: In this dialogue, the analysis has identified sentiments and rationales behind opinions, including $h_1$'s opinion $o_1$ on $a_1$ of $t_1$ carries a sentiment $s_1$ with rationale $r_1$, etc. 
Please based on the dialogue, verify whether these descriptions are consistent with the dialogue content and provide `1' for `yes' or `0' for `no' judgment.\\

\textbf{\color{blue}{Expected Output}}: 1 (if yes) or 0 (if no)}
\end{tcolorbox}

\paratitle{\(\blacktriangleright\) Step 4: Verification of Sentiment Flipping Trigger Classification}
\begin{tcolorbox}[breakable, fontupper=\customfont]
{\small
\textbf{Input Data}: \(D\), \(\{(h_j, t_i, a_i, o_j, s_k, r_l)\}\) \\
\textbf{Instruction}: In this dialogue, instances of sentiment flipping and their triggers have been identified, including $h_1$'s sentiment towards $a_1$ of $t_1$ initially was $\zeta_1$ and later flipped to $\phi_1$ due to trigger $\tau_1$, etc. 
Please based on the dialogue and your commonsense knowledge, verify whether these descriptions accurately capture the emotional dynamics and their triggers in the dialogue and provide `1' for `yes' or `0' for `no' judgment. \\

\textbf{\color{blue}{Expected Output}}: 1 (if yes) or 0 (if no)}
\end{tcolorbox}

Upon receiving outcomes from the verification prompted by the MLLM, the next steps are as follows:

\textbf{In case of inconsistency}: If verification results show the expression is inconsistent with the dialogue content, we will instruct the LLM to regenerate and reverify the k-tuples.

\textbf{In case of consistency}: If the LLM confirms the expression is consistent with the dialogue content, it indicates that the current step's reasoning and transformation results are trustworthy. We then proceed with the next steps of analysis and verification based on this confirmed information.

This procedure ensures the analysis moves forward in an orderly manner. 
If inconsistencies arise, they are addressed by revisiting the analysis steps; once results are confirmed to be consistent, the analysis proceeds, leveraging these verified outcomes for subsequent steps.

\section{Extensions of Settings and Implementations}


In this section, we continue to provide more descriptions about the implementation details of our system and experiments.

\subsection{System Training Details}


\paratitle{E.1.1 Training Step 1: Multimodal Understanding Stage}

\paratitle{\(\blacktriangleright\) Training Data}: The training data comprises `text+X' pairs, where `X' represents various forms of multimodal inputs including images, audios or videos. 
This diverse dataset structure is crucial for enabling LLM to learn from and interpret a wide range of multimodal information, thereby enhancing its ability to process and understand complex multimodal scenarios. 
Specifically, we employ well-established datasets such as LLaVA\cite{liu2024visual}, miniGPT-4\cite{zhu2023minigpt} and VideoChat\cite{li2023videochat}, which have been designed for multimodal language model instruction tuning. 
These datasets not only provide a rich source of `Text+X' pairs but also align with our objective to improve LLM's proficiency in generating textual responses from multimodal inputs, covering a broad spectrum of real-world scenarios and enhancing the model's understanding of multimodal content.

\paratitle{\(\blacktriangleright\) Training Objective:} The primary objective is to train the LLM to accurately interpret and generate textual descriptions for multimodal inputs, fostering a comprehensive understanding of both textual and non-textual content.

\paratitle{\(\blacktriangleright\) Training Method:} The multimodal inputs are encoded using the ImageBind model. The encoded information is then projected into the model's embedding space using a projection layer. The language model is fine-tuned using LoRA.

\paratitle{\(\blacktriangleright\) Loss Function}: To optimize the model's ability to generate accurate textual descriptions from multimodal inputs, we employ the Negative Log-Likelihood (NLL) Loss.
\begin{equation}
L_{NLL} = -\sum_{t=1}^{T} \log(p_{t,c_t})
\end{equation}
where \(T\) is the length of the text sequence, \(c_t\) represents the correct word at time step \(t\), and \(p_{t,c_t}\) is the probability assigned by the model to the correct word at time step \(t\).

\paratitle{E.1.2 Training Step 2: Sextuple Extraction Understanding}

\paratitle{\(\blacktriangleright\) Training Data}: Use the PanoSent train set as supervised data, containing input dialogues and the corresponding sextuple information extracted from these dialogues.

\paratitle{\(\blacktriangleright\) Training Objective}: To train the model to accurately understand and extract sextuple information from dialogues, mastering the specified input-output pattern.

\paratitle{\(\blacktriangleright\) Training Method:} The base model is fine-tuned using LoRA, fitting the model to predict the correct sextuple information based on the input dialogues.

\paratitle{\(\blacktriangleright\) Loss Function}: To optimize the model's performance in accurately extracting sextuple information from dialogues, we also apply the Negative Log-Likelihood (NLL) Loss.
\begin{equation}
L_{NLL} = -\sum_{t=1}^{T} \log(p_{t,c_t})
\end{equation}
where \(T\) is the length of the output sequence, \(c_t\) represents the correct label at time step \(t\), and \(p_{t,c_t}\) is the probability assigned by the model to the correct label at time step \(t\).

\paratitle{E.1.3 Training Step 3: Paraphrase-based Verification}

\paratitle{\(\blacktriangleright\) Training Data:} Comprises paraphrase pairs that exhibit either an entailment or contradiction relation to the given context, aimed at verifying the accuracy of results from previous reasoning steps.

\paratitle{\(\blacktriangleright\) Training Objective:} To train the model to distinguish between entailment and contradiction in the context of the provided paraphrases, ensuring the integrity and reliability of each reasoning step.

\paratitle{\(\blacktriangleright\) Loss Function:} For the task of classifying paraphrase pairs as entailment or contradiction, we use the Binary Cross-Entropy Loss function.
\begin{equation}
L_{BCE} = -\frac{1}{N} \sum_{i=1}^{N} \left[ y_i \log(p_i) + (1 - y_i) \log(1 - p_i) \right]
\end{equation}
where \(N\) is the number of samples, \(y_i\) indicates the true label (1 for entailment, 0 for contradiction), and \(p_i\) is the predicted probability of the \(i^{th}\) sample being an entailment. This loss function aims to optimize the model's ability to accurately classify the paraphrase pairs into the correct categories, enhancing the accuracy of the reasoning process.

\section{Evaluation Specifications}
\label{Evaluation Specifications}

Here, we provide a detailed introduction on how we conduct the evaluation for the two subtasks.

\subsection{Subtask-I Evaluation}

\label{Evaluation Specifications1}

For Subtask I, focusing on the extraction of fine-grained sentiment sextuples, our evaluation methodology is designed to rigorously assess the performance across various aspects of the task. We provide detailed specifications for element-wise, pair-wise, and overall sextuple evaluations.

\paratitle{F.1.1 Element-wise Evaluations}

\paratitle{\(\blacktriangleright\) Explicit Elements.} 
For elements explicitly mentioned in the text, we apply the exact match metric for evaluation. 
  Under this metric, a correct prediction must precisely match the term as annotated in the gold standard. 
  Exact Precision (EP) is calculated as the proportion of correctly predicted terms among all predicted terms, while Exact Recall (ER) is the proportion of correctly predicted terms among all gold terms. 
\begin{equation}
  EP = \frac{\text{\#correct terms}}{\text{\#predicted terms}}
  \label{eq:EP}
\end{equation}
\begin{equation}
  ER = \frac{\text{\#correct terms}}{\text{\#gold terms}}
  \label{eq:ER}
\end{equation}
\begin{equation}
  \text{Exact Match F1} = 2 \cdot \frac{EP \cdot ER}{EP + ER}
  \label{eq:Exact F1}
\end{equation}
Here, `\#’ denotes the amount, and `correct terms’ refer to the predicted terms that exactly match the gold terms.

\paratitle{\(\blacktriangleright\) Implicit Elements.} For implicit elements not explicitly mentioned in the text, we utilize the binary match metric, which is a relaxation of the above exact one.
    For implicit elements not explicitly mentioned in the text, we utilize the binary match metric, which is a relaxation of the exact match metric. We evaluate if the predicted element is semantically identical to the gold term, as assessed by GPT-4, assigning a binary outcome (1 if yes, otherwise 0). When constructing such queries, it is crucial to include sufficient contextual information from the dialogue. This is because the meaning of terms can vary with context, and relying solely on the terms themselves may not accurately reflect their significance in a specific dialogue. Therefore, it is essential that the prompts provided to GPT-4 contain complete dialogue content to enable accurate semantic evaluation. Our standard instruction template for GPT-4 is: "Given the context of the dialogue, do `[predicted term]' and `[gold standard term]' have similar meanings?"
\begin{equation}
  BP = \frac{\text{\#semantically identical terms}}{\text{\#predicted terms}}
  \label{eq:BP}
\end{equation}
\begin{equation}
  BR = \frac{\text{\#semantically identical terms}}{\text{\#gold terms}}
  \label{eq:BR}
\end{equation}
\begin{equation}
  \text{Binary Match F1} = 2 \cdot \frac{BP \cdot BR}{BP + BR}
  \label{eq:Binary F1}
\end{equation}

\paratitle{\(\blacktriangleright\) Element of Explicit Rationale.} For evaluating the explicit rationale element, we use the proportional match metric, which measures the proportional overlap between the predicted and gold standard terms.
  Proportional overlap assigns a score to represent the proportion of the overlapped region, rather than a binary value, 0 or 1.
  Proportional precision (PP) measures the proportion of the overlap between a predicted term and an overlapping gold term. 
  Proportional recall (PR) measures the proportion of the overlap between a gold term and an overlapping predicted term. 

  \begin{equation}
  PP = \frac{\text{\#correct terms\textbar proportional overlap}}{\text{\#predicted terms}}
  \label{eq:PP}
\end{equation}
\begin{equation}
  PR = \frac{\text{\#correct terms\textbar proportional overlap}}{\text{\#gold terms}}
  \label{eq:PR}
\end{equation}
\begin{equation}
  \text{Proportional Match F1} = 2 \cdot \frac{PP \cdot PR}{PP + PR}
  \label{eq:Proportional F1}
\end{equation}

\paratitle{\(\blacktriangleright\) F1 Score for Each Element.} The F1 score for each element is the average of the Exact F1 and the Relevant F1 score under that category, which could be either Binary F1 for implicit elements or Proportional F1 for explicit rationale, depending on the nature of the element.

\paratitle{\(\blacktriangleright\) Sentiment Classification.}
The macro F1 Score is calculated as the average of F1 Scores for all sentiment classes, offering a balanced measure of model performance across different sentiment orientations. For each sentiment class $c$, we define:

\begin{equation}
  CP_c = \frac{\text{\#correct predictions for class } c}{\text{\#predictions of class } c}
\end{equation}

\begin{equation}
  CR_c = \frac{\text{\#correct predictions for class } c}{\text{\#gold instances of class } c}
\end{equation}

\begin{equation}
  \text{Class F1}_c = 2 \cdot \frac{CP_c \times CR_c}{CP_c + CR_c}
\end{equation}

\begin{equation}
\text{Macro F1} = \frac{\text{F1}_{\text{positive}} + \text{F1}_{\text{negative}} + \text{F1}_{\text{neutral}}}{3}
\end{equation}

\paratitle{F.1.2 Pair-wise Evaluations}

For a pair, the prediction must correctly identify both spans, and adhere to the evaluation standards for implicit elements and rationale. 

\paratitle{\(\blacktriangleright\) Pair-wise F1 Score.} This metric evaluates the precision and recall of correctly identified pairs within the sextuples.
\begin{equation}
  PP = \frac{\text{\#correct pairs}}{\text{\#predicted pairs}}
\end{equation}
\begin{equation}
  PR = \frac{\text{\#correct pairs}}{\text{\#gold pairs}}
\end{equation}
\begin{equation}
  \text{Pair-wise F1} = 2 \cdot \frac{PP \cdot PR}{PP + PR}
\end{equation}

\paratitle{F.1.3 Sextuple Evaluations}

For sextuple extraction, the prediction must accurately match all six elements, samely with consideration for the accuracy of implicit elements and rationale. 

\paratitle{\(\blacktriangleright\) Micro F1 Score.} This metric evaluates the overall precision(OP) and overall recall(OR) for sextuple extraction.
\begin{equation}
  OP = \frac{\text{\#correct sextuples}}{\text{\#predicted sextuples}}
\end{equation}
\begin{equation}
  OR = \frac{\text{\#correct sextuples}}{\text{\#gold sextuples}}
\end{equation}
\begin{equation}
  \text{Micro F1} = 2 \cdot \frac{OP \cdot OR}{OP + OR}
\end{equation}

\paratitle{\(\blacktriangleright\) Identification F1 Score.} This metric focuses on the identification precision(IP) and identification recall(IR) of sextuples, excluding sentiment polarity.
\begin{equation}
  IP = \frac{\text{\#correctly identified sextuples without sentiment}}{\text{\#predicted sextuples}}
\end{equation}
\begin{equation}
  IR = \frac{\text{\#correctly identified sextuples without sentiment}}{\text{\#gold sextuples}}
\end{equation}
\begin{equation}
  \text{Identification F1} = 2 \cdot \frac{IP \cdot IR}{IP + IR}
\end{equation}

\subsection{Subtask-II Evaluation}

In Subtask-II, the evaluation of model performance in identifying sentiment flips and their triggers adopts specific measures tailored to the complexity of each task component. 
For assessing the identification of initial and flipped sentiments as well as their combined evaluation with triggers, the exact match F1 score is employed to account for the precision in capturing the interconnected aspects of sentiment transitions. 
Conversely, for the classification task of identifying triggers alone, the Macro F1 score is utilized to ensure a balanced evaluation across all trigger categories, reflecting equal importance to the accurate identification of each trigger type.

\paratitle{F.2.1 Flip Evaluations}

To assess the model's ability to correctly identify both the initial sentiment and the flipped sentiment, we use the exact match F1 score. This measure accurately reflects the model's capability in detecting precise changes in sentiment:
\begin{equation}
\text{Exact Match F1} = 2 \cdot \frac{\text{Precision}_{\text{Flip}} \times \text{Recall}_{\text{Flip}}}{\text{Precision}_{\text{Flip}} + \text{Recall}_{\text{Flip}}}
\end{equation}

\paratitle{F.2.2 Trigger Evaluations}

We evaluate the identification of flipping triggers using the Macro F1 score, which accommodates the diversity of trigger categories within the dataset. This metric ensures that all categories are assessed with equal importance, providing a balanced measure of performance across varied types of triggers.
\begin{equation}
\text{Macro F1} = \frac{1}{N} \sum_{i=1}^{N} 2 \cdot \frac{\text{Precision}_{i} \times \text{Recall}_{i}}{\text{Precision}_{i} + \text{Recall}_{i}}
\end{equation}
where $N$ is the number of trigger categories, and $\text{Precision}_{i}$ and $\text{Recall}_{i}$ are the precision and recall for the $i$-th trigger category, respectively.

\paratitle{F.2.3 Overall Flip-Trig Evaluations}

Finally, the model's overall performance in simultaneously identifying both the correct flipped sentiment and the correct trigger is assessed using the exact match f1 score, providing a comprehensive evaluation of the model's nuanced understanding of sentiment dynamics and their triggers:
\begin{equation}
\text{Exact Match F1} = 2 \cdot \frac{\text{Precision}_{\text{Flip-Trig}} \times \text{Recall}_{\text{Flip-Trig}}}{\text{Precision}_{\text{Flip-Trig}} + \text{Recall}_{\text{Flip-Trig}}}
\end{equation}

\section{More Experiments and Analyses}

We further present additional experimental results and analyses.

\subsection{Evaluation on Rationale}

This experiment aims to compare the applicability of the proportional match F1 versus exact match F1 evaluation metrics in the task of rationale extraction. 
We focus on empirically validating the performance of these two evaluation methods across 200 data entries, using human judgment as a benchmark to assess their effectiveness.

First, we calculate the exact match F1 and proportional match F1 scores for rationale extraction on the selected dataset. 
Next, we conduct a manual review of these 200 data entries, providing a binary match F1 score to assess whether the predicted rationale is semantically identical to the gold rationale. 
Lastly, these automatically computed scores are directly compared with the results of the manual review.

As shown in Table~\ref{proportionalf1}, the results demonstrate that the proportional match F1 scores are significantly more consistent with manual evaluations than the exact match F1 scores. 
This finding supports the effectiveness of the proportional match F1 evaluation metric in situations of partial text match for rationale. It indicates that proportional match F1 better captures and evaluates text segments that support specific sentiment judgments, compared to exact match F1. 
This discrepancy highlights the superior flexibility and alignment of proportional match F1 with human assessment practices in sentiment analysis tasks, especially those involving rationale extraction.

\begin{table}[!t]
\fontsize{8}{9.5}\selectfont
\setlength{\tabcolsep}{2mm}
\centering
\caption{Rationale extraction evaluation results on 200 EN test samples.}
\vspace{-3mm}
\begin{tabular}{lccc}
\toprule
& \bf Rationale Extraction \\
\midrule
Human Evaluation & 67.31\\
Proportional Match F1 & \textbf{45.49} \\
Exact Match F1 & 20.38 \\ 
\bottomrule
\end{tabular}
\vspace{-3mm}
\label{proportionalf1}
\end{table}

\subsection{Extended Explorations of Impact of Using Different Backbone LLMs}

In order to compare the performance of different LLM backbones on our two subtasks, we conduct a controlled experiment where we maintain consistent methodologies and architectures across two settings—Sentica and Sentica (+CoS+PpV)—while varying only the LLM backbone used for task reasoning.
For a fair comparison, each model is evaluated using the same set of parameters and input data (only English dataset), ensuring that any performance differences could be attributed to the backbone itself, rather than external variables.

As presented in Table~\ref{diff_backbone}, the results indicate that the Flan-T5-XXL backbone outperforms others in both subtasks. 
This superior performance is evident in the consistently higher scores achieved in the subtasks, confirming the efficacy of Flan-T5-XXL as a backbone for the Sentica framework.

 \begin{table}[!t]
\fontsize{8}{9.5}\selectfont
\setlength{\tabcolsep}{1.7mm}
\centering
\caption{Comparason of LLM Backbones on EN Dataset.}

\vspace{-3mm}
\begin{tabular}{@{} llllccccccccc @{}}
\toprule
&\multirow{2}{*}{\textbf{PLM}} & \multirow{2}{*}{\textbf{Method}} & \multicolumn{2}{c}{\textbf{Result}} \\
\cmidrule(lr){4-6}
&& & Sextuple & Flip-Trig \\
\midrule
M1 & Llama 2 & Sentica & 19.58 & 55.46 \\
M2 & Llama 2 & Sentica(+CoS+PpV) & 28.70 & 65.63 \\
\hdashline
M3 & Vicuna 7B & Sentica & 20.16 & 56.09 \\
M4 & Vicuna 7B & Sentica(+CoS+PpV) & 29.97 & 67.03 \\
\hdashline
M5 & Flan-T5-XXL & Sentica & 21.26 & 58.45 \\
M6 & Flan-T5-XXL & Sentica(+CoS+PpV) & \textbf{32.18} & \textbf{69.39} \\
\bottomrule
\end{tabular}
\vspace{-2mm}
\label{diff_backbone}
\end{table}


\subsection{Cross-utterance Sextuple Extraction and Sentiment Flip Trigger Identification.} 
\begin{figure}[!t]
\includegraphics[width=0.98\columnwidth]{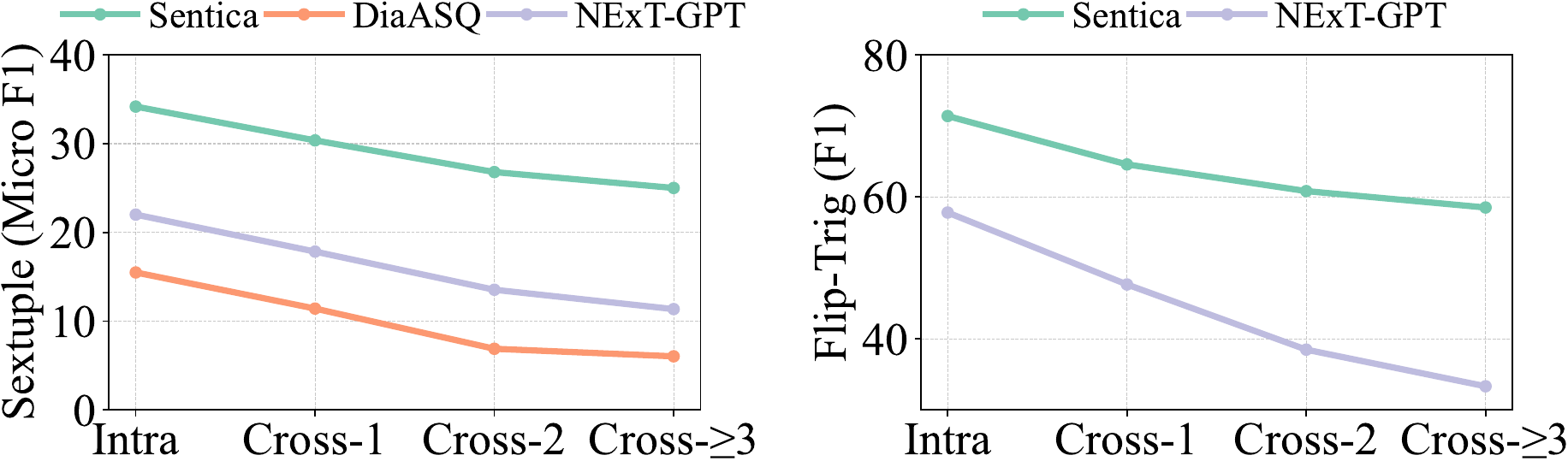}
\caption{
Performance of two subtasks on different cross-utterance levels.
}
\label{crossutterance}
\vspace{-1mm}
\end{figure}

In assessing the impact of cross-utterance dialogue dynamics on emotion analysis tasks, our experimental results demonstrate a consistent trend across both subtasks evaluated, shown in Figure~\ref{crossutterance}. 
Cross-utterance interaction presents a discernible challenge that invariably leads to a degradation in performance. 
However, our Sentica mitigates this effect more robustly than comparative methodologies. 
This is evidenced by a relatively smaller decline in F1 scores, particularly in scenarios with increased cross-utterance complexity. 
Subtask I, which entails the extraction of sentiment sextuples, inherently requires a deeper contextual comprehension, making it more vulnerable to cross-utterance disturbances than Subtask II's focus on sentiment trigger identification and classification. 
When comparing LLM-based methods (Sentica and NExT-GPT) with non-LLM-based methods (DiaASQ), the former exhibits superior capability in contending with cross-utterance intricacies. Specifically, our model outstrips DiaASQ significantly under cross-utterance conditions, maintaining a higher performance baseline. 
For Subtask II, a similar pattern prevails with our method outperforming NExT-GPT. 
This underlines our model's robustness, not only in intra-utterance contexts but also when navigating the complexities introduced by cross-utterance dialogue sequences.

\subsection{Impact of Joint VS. Separate Subtask Execution.}

The experiment aims to determine the effects of jointly performing Panoptic Sentiment Sextuple Extraction (subtask-I) and Sentiment Flipping Analysis (subtask-II) as opposed to processing them separately. 
In our CoS framework, we adopt a joint (cascade) approach.
Comparative analysis of the results reveals that Subtask-II, when informed by the sentiment sextuples inferred from Subtask-I, demonstrates increased accuracy in identifying the Flip-Tri pair within dialogues. 
This improvement is significantly reflected in the increase of the Flip-Tri pair metric from 60.06 to 69.39, as shown in Table~\ref{tab:comparison}. 
The findings confirm that the sentiment sextuples from Subtask-I serve as critical reference information for Subtask-II, significantly enhancing the precision of sentiment flip identification and analysis, thereby highlighting the necessity and efficacy of an integrated approach to complex sentiment analysis tasks.

\begin{table}[!t]
\fontsize{8}{9.5}\selectfont
\setlength{\tabcolsep}{2.5mm} 
\centering
\caption{Comparison of joint and separate execution for subtask-II on EN data.}
\begin{tabular}{@{} lcc @{}}
\toprule
& \bf Sextuple & \bf Flip-Trig Trip \\
\midrule
Joint & 32.18 & \textbf{69.39} \\
Separate & 32.18 & 60.06 \\
\bottomrule
\end{tabular}
\label{tab:comparison}
\end{table}


\subsection{Influence of Training with Different Data Amount}

\begin{figure}[!t]
 \centering
 \includegraphics[width=\columnwidth]{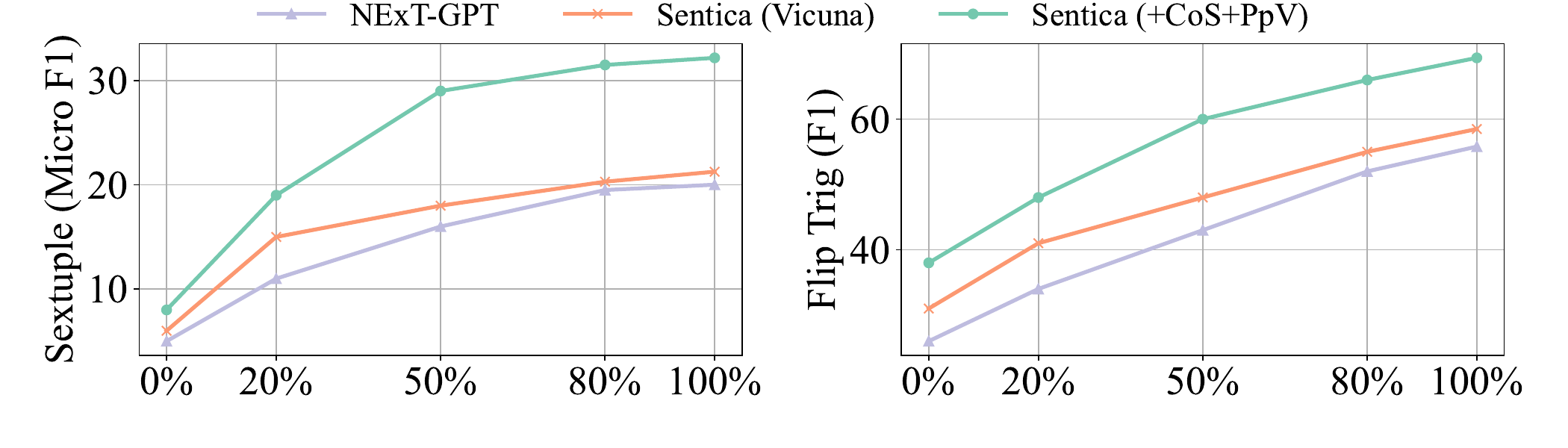}
 \caption{Performance on different training data volume.}
 \label{trainingvolume}
 \vspace{-1mm}
\end{figure}

In this study, we explore the effects of varying the volume of supervised training data on a LLM across five different data levels: 0\%, 20\%, 50\%, 80\%, and 100\% of the training set. 
This investigation aims to pinpoint how different quantities of training data influence the model's performance in a supervised setting, with a particular focus on understanding the incremental benefits of additional data.
We systematically increase the proportion of the dataset used for training, allowing for a direct comparison of the model's performance across these varying levels of data availability. 
The result, as shown in Figure~\ref{trainingvolume}, shows a consistent improvement in the model's effectiveness as the amount of supervised training data increases. Notably, the increase from 0\% to 20\% of the training data yields the most significant performance boost, demonstrating that early additions of supervised data substantially enhance the model's capabilities.

\subsection{Few-shot Learning Experiments}

\begin{figure}[!t]
 \centering
 \includegraphics[width=\columnwidth]{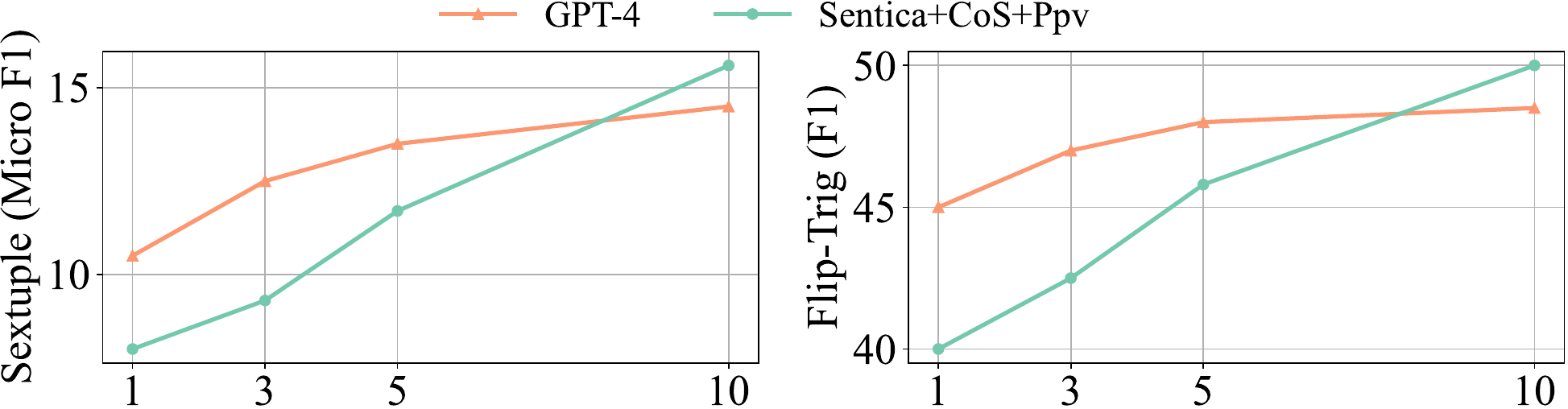}
 \caption{Performance comparison of our model and GPT-4 across different few-shot learning.}
 \label{fewshot}
 \vspace{-1mm}
\end{figure}

The experiment is designed to compare the efficacy of our model against GPT-4 in few-shot learning scenarios without prior task-specific training. 
In conducting this comparison, few-shot instances of 1, 3, 5, and 10 are chosen to observe how both models adapt and learn from an increasing number of examples. The results in Figure~\ref{fewshot} shows that both models performing modestly with just 1 and 3 shots, due to the limited amount of information available. GPT-4 performs significantly better in scenarios with minimal examples.
However, as the shot count is elevated to 5 and then to 10, our model demonstrate a notable uptick in performance, indicative of its enhanced capability to assimilate and apply the task's salient features and patterns effectively.

\subsection{Case Study}

\begin{figure*}[!t]
 \centering
 \includegraphics[width=0.96\linewidth]{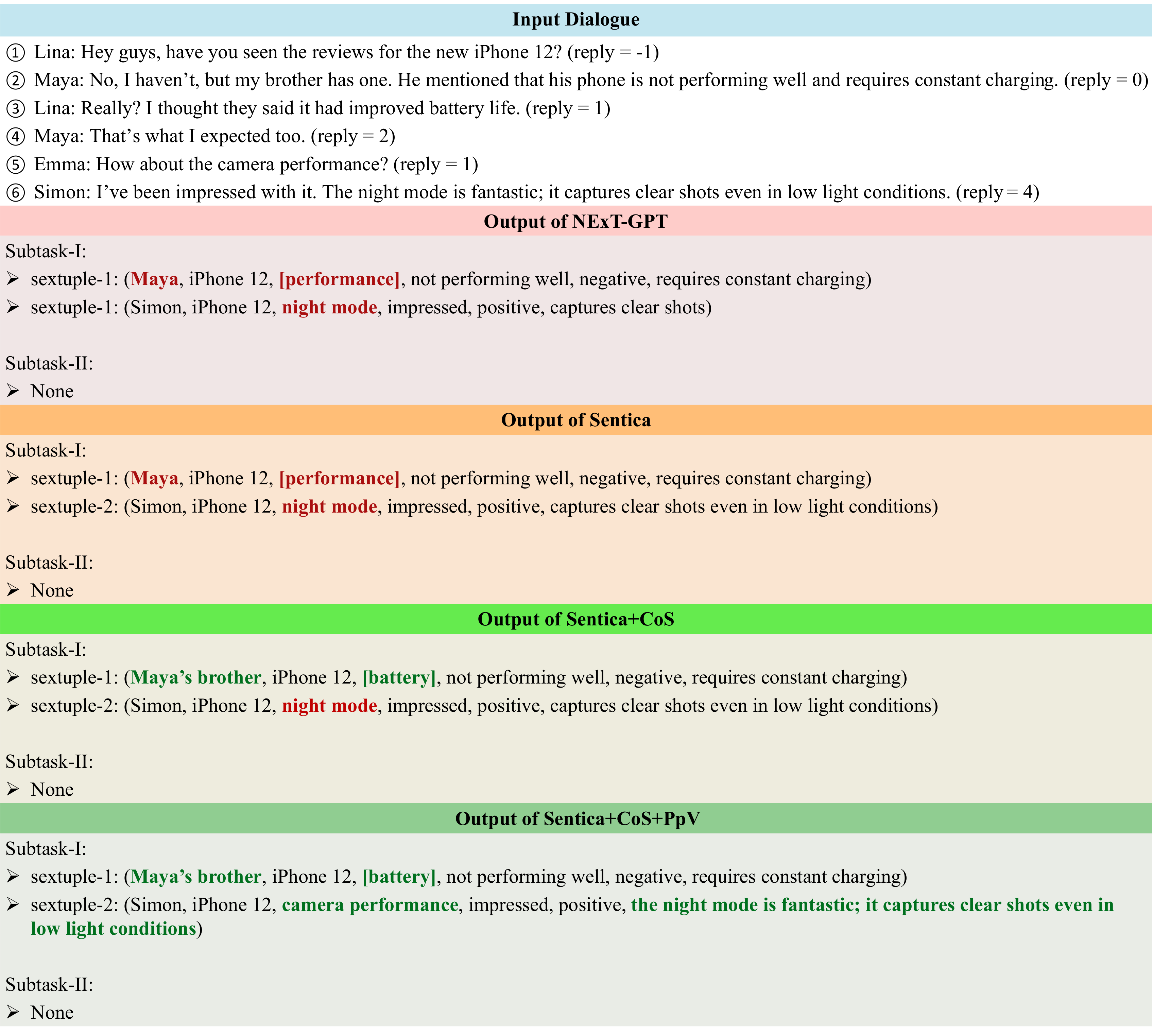}
 \caption{A conversation on domain of electronic products. Different colors represent two types of answers. The first type in red indicates the wrong one, yet the green is the correct answer. \texttt{[*]} indicates the implicit information in the text.}
 \label{case1}
  \vspace{-1mm}
\end{figure*}

\begin{figure*}[!t]
 \centering
 \includegraphics[width=1.0\linewidth]{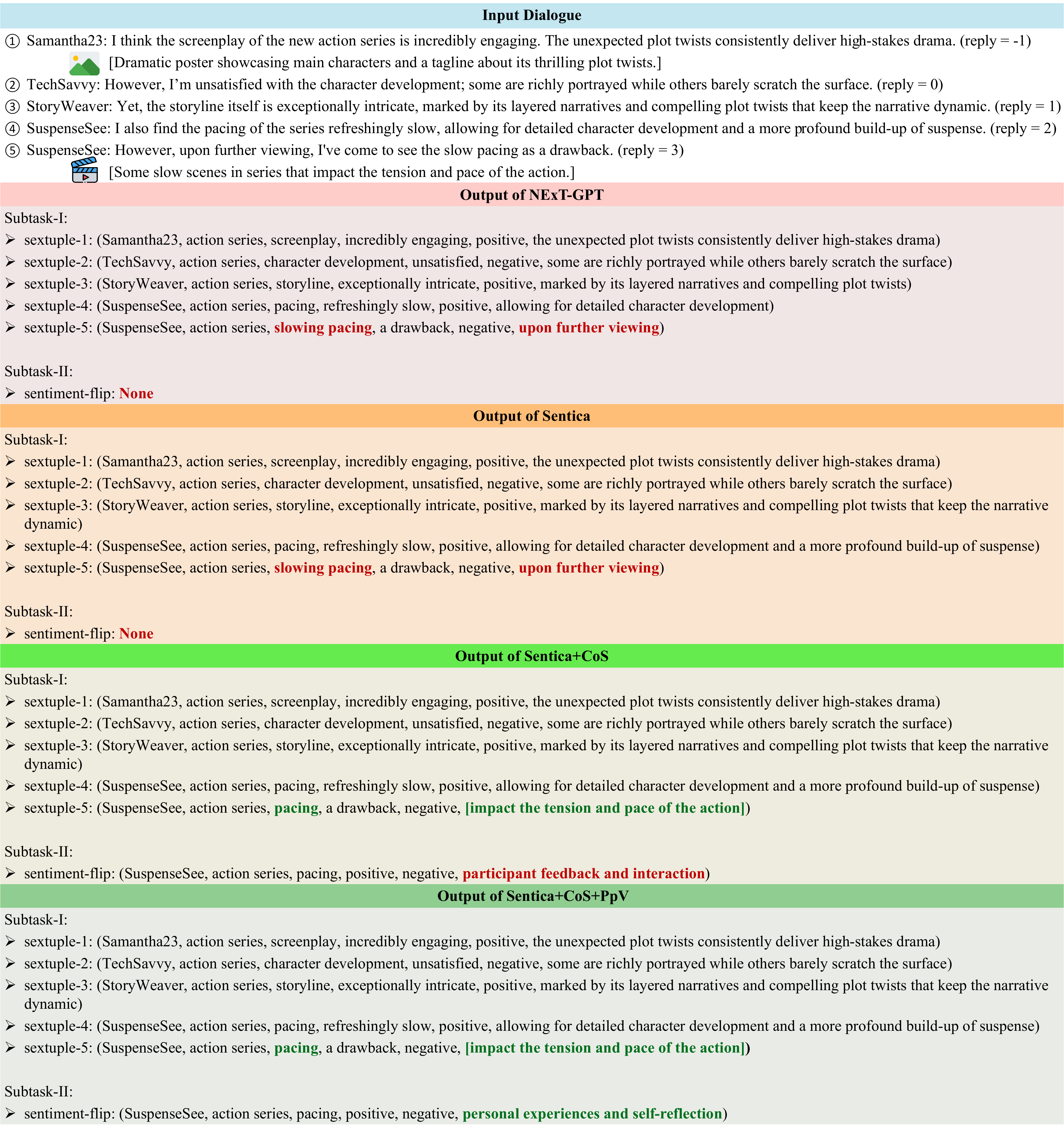}
 \caption{A conversation on domain of movies and entertainment.}
 \label{case2}
  \vspace{-1mm}
\end{figure*}

\begin{figure*}[!t]
 \centering
 \includegraphics[width=0.875\linewidth]{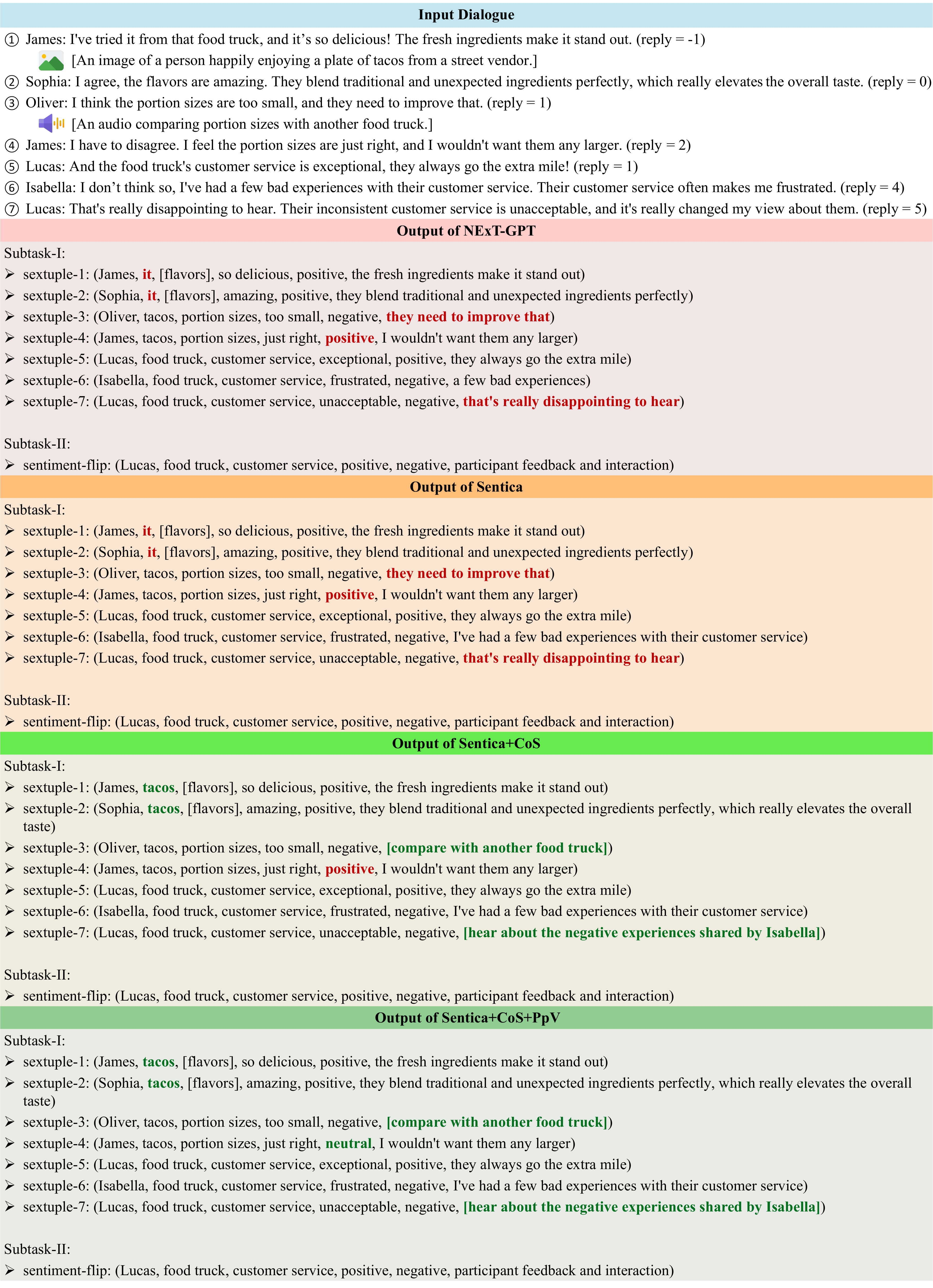}
 \caption{A conversation on domain of food and cuisine.}
 \label{case3}
  \vspace{-1mm}
\end{figure*}

We present several examples to highlight the performance differences between our model and others. 
As shown in Figures~\ref{case1}, \ref{case2}, and \ref{case3}, our model exhibits a deeper understanding of complex dialogue contexts, skillfully capturing subtle nuances and inferring implicit intentions. 
Its superior ability to handle multimodal information results in a more accurate interpretation across various modalities. Additionally, our model excels at uncovering implicit elements within dialogues. 
These strengths collectively allow for more comprehensive extraction of sextuple information and also aid in a more accurate analysis of sentiment flips within conversations.

\end{document}